%% file: template.tex
\documentclass{article}

\usepackage{arxiv}

\usepackage[utf8]{inputenc} 
\usepackage[T1]{fontenc}    
\PassOptionsToPackage{hyphens}{url}
\usepackage{hyperref}       
\usepackage{amsfonts}       
\usepackage{nicefrac}       
\usepackage{microtype}      
\usepackage{lipsum}		
\usepackage{graphicx}
\usepackage{natbib}
\usepackage{doi}
\usepackage[hyphens]{url}  
\usepackage{caption} 
\usepackage{algorithm}
\usepackage{algorithmic}

\usepackage{booktabs}
\usepackage{makecell}
\usepackage{multirow}
\usepackage{placeins}
\usepackage{rotating}
\usepackage{xcolor}
\definecolor{MUEHeatOne}{HTML}{F4F7FA}
\definecolor{MUEHeatTwo}{HTML}{E6EDF3}
\definecolor{MUEHeatThree}{HTML}{D2DFEA}
\definecolor{MUEHeatFour}{HTML}{BCD0E0}
\definecolor{MUEHeatFive}{HTML}{9FB9D0}
\usepackage{colortbl}

\usepackage{newfloat}
\usepackage{listings}

\usepackage{booktabs}

\title{MUSE: Benchmarking Large Vision-Language Models on Multi-Modal
Understanding in Situated Education}

\newcommand{\museaffiliations}{}
\newcommand{\affiliations}[1]{\renewcommand{\museaffiliations}{#1}}

\author{
    \begin{minipage}[t]{0.95\textwidth}
    \centering
    \bfseries
    \href{https://orcid.org/0000-0002-7422-7318}{\includegraphics[scale=0.06]{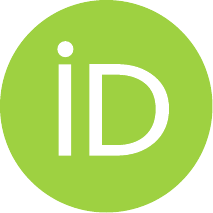}\hspace{1mm}Luyao~Zhu\textsuperscript{\textrm{1}}},
    \hspace{1mm}Xun Wei~Yee\textsuperscript{\textrm{1}},
    \href{https://orcid.org/0000-0002-8077-7025}{\includegraphics[scale=0.06]{orcid.pdf}\hspace{1mm}Wei~Li\textsuperscript{\textrm{3}}}\thanks{Corresponding author.},
    \hspace{1mm}Mun Thye~Mak\textsuperscript{\textrm{1}},
    \href{https://orcid.org/0000-0003-3523-3210}{\includegraphics[scale=0.06]{orcid.pdf}\hspace{1mm}Wee Siong~Ng\textsuperscript{\textrm{2}}}
    \\[0.75em]
    \normalfont
    \museaffiliations
    \end{minipage}
}
\affiliations{
    \textsuperscript{\rm 1} AI Singapore, National University of Singapore, Singapore\\
    \textsuperscript{\rm 2} School of Computing, National University of Singapore, Singapore\\
    \textsuperscript{\rm 3} Institute of Advanced Intelligence and Computing, A*STAR\\[0.5em]
    Luyao Zhu: \href{mailto:luyaozhu@outlook.com}{luyaozhu@outlook.com}\\
    Wei Li: \href{mailto:wei008@e.ntu.edu.sg}{wei008@e.ntu.edu.sg}\\
    Wee Siong Ng:
    \href{mailto:Ng_Wee_Siong@a-star.edu.sg}{Ng\_Wee\_Siong@a-star.edu.sg}
}

\renewcommand{\shorttitle}{\textit{arXiv} Template}

\hypersetup{
pdftitle={A template for the arxiv style},
pdfsubject={q-bio.NC, q-bio.QM},
pdfauthor={David S.~Hippocampus, Elias D.~Striatum},
pdfkeywords={First keyword, Second keyword, More},
}

\begin{document}
\maketitle

\begin{abstract}
	Large vision-language models have achieved remarkable progress in multi-modal understanding, yet their capabilities in educational settings remain insufficiently evaluated. In AI-assisted language learning, models must interpret artistic imagery, understand its semantic, affective, and cultural content, and reason about visual context to support meaningful interaction. However, existing benchmarks primarily focus on real-world images or domain-specific educational reasoning, providing limited coverage of artistic educational content. To address this gap, we introduce MUSE, a benchmark for evaluating large vision-language models on artistic image understanding in situated educational applications. MUSE decouples image annotation from question generation, enabling diverse tasks with controllable difficulty while reducing annotation effort. It comprises twelve tasks spanning visual perception, semantic and affective interpretation, culture understanding, and compositional reasoning, together with diverse artistic images deliberately curated to center Singaporean and Southeast Asian multicultural contexts alongside Western art traditions, covering multiple themes and difficulty levels. Evaluation of open-source and proprietary models reveals substantial disparities across capability dimensions, particularly in affective interpretation and compositional reasoning. Our analysis further identifies common failure modes and key challenges for developing trustworthy multi-modal models for education. We hope MUSE will serve as a standardized benchmark for advancing multi-modal understanding in situated educational applications.
\end{abstract}


\begin{center}
    \small
    \hypersetup{hidelinks}
    \href{https://huggingface.co/datasets/Cyn7hia-Z/MUSE/tree/main/code}
         {\textcolor{blue!60!black}{\textbf{Code}}}
    \quad\textbar\quad
    \href{https://huggingface.co/datasets/Cyn7hia-Z/MUSE}
         {\textcolor{blue!60!black}{\textbf{Dataset}}}
\end{center}

\section{Introduction}
Large vision-language models (VLMs) have made substantial progress in integrating visual perception with language understanding and generation, enabling tasks such as visual question answering, image description, visual grounding, multi-modal dialogue, and visual reasoning~\cite{openai2023gpt4v,bai2023qwenvlversatilevisionlanguagemodel,chen2024internvl}. Their growing capabilities have encouraged applications in situated education, including intelligent tutoring, personalized learning, automated feedback, and multi-modal content interaction~\cite{chu2025uniedu}. A central requirement in these settings is the ability to interpret instructional images and connect their visual content with meaningful linguistic representations. This is especially demanding in image-based learning, where artworks prompt vocabulary use, description, narrative construction, emotional expression, and cultural discussion~\cite{zhuang2024visual,shimabukuro-etal-2025-langeye}. An AI tutor must interpret the same image to formulate questions, assess responses, explain linguistic concepts, and provide appropriate feedback, requiring semantic, affective, spatial, compositional, and cultural understanding beyond object recognition.
\begin{table}[t]
\centering
\footnotesize
\begin{tabular}{@{}>{\raggedright\arraybackslash}p{0.21\linewidth}>{\raggedright\arraybackslash}p{0.35\linewidth}>{\raggedright\arraybackslash}p{0.40\linewidth}@{}}
\toprule
\textbf{Dimension} & \textbf{Capability} & \textbf{Tasks} \\
\midrule

Visual Perception &
Objects and their quantities &
Object Classification, Object Count \\
\midrule

Semantic Understanding &
Scenes, human activities, and events &
Scene Classification, Activity Localization, Activity Description \\
\midrule

Affective Interpretation &
Emotions, their causes, and supporting visual evidence &
Emotion Detection, Emotion Cause Inference, Visual Clue Identification \\
\midrule

Compositional Reasoning &
Spatial and structural composition &
Relative Position, Remote Interaction, Jigsaw Puzzle \\
\midrule

Cultural Understanding &
Cultural-specific visual knowledge &
Cultural Identification \\
\bottomrule
\end{tabular}
\caption{Capability dimensions in MUSE.}
\label{tab:capability_dimensions}
\end{table}

Artistic imagery, such as paintings, illustrations, and cartoons, further complicates this task. Compared with natural photographs, these images often contain stylized or exaggerated forms, non-photorealistic colors, implicit narratives, and culturally dependent cues. General-purpose VLMs have shown limitations in interpreting such content, motivating dedicated models and benchmarks for artistic understanding~\cite{yuan2023artgpt4,alfarano2025vqartbench}. Consequently, performance on natural-image benchmarks may not reliably reflect a model's ability to understand artistic imagery in educational settings.

\begin{figure*}[t]
    \centering
    \includegraphics[width=\textwidth]{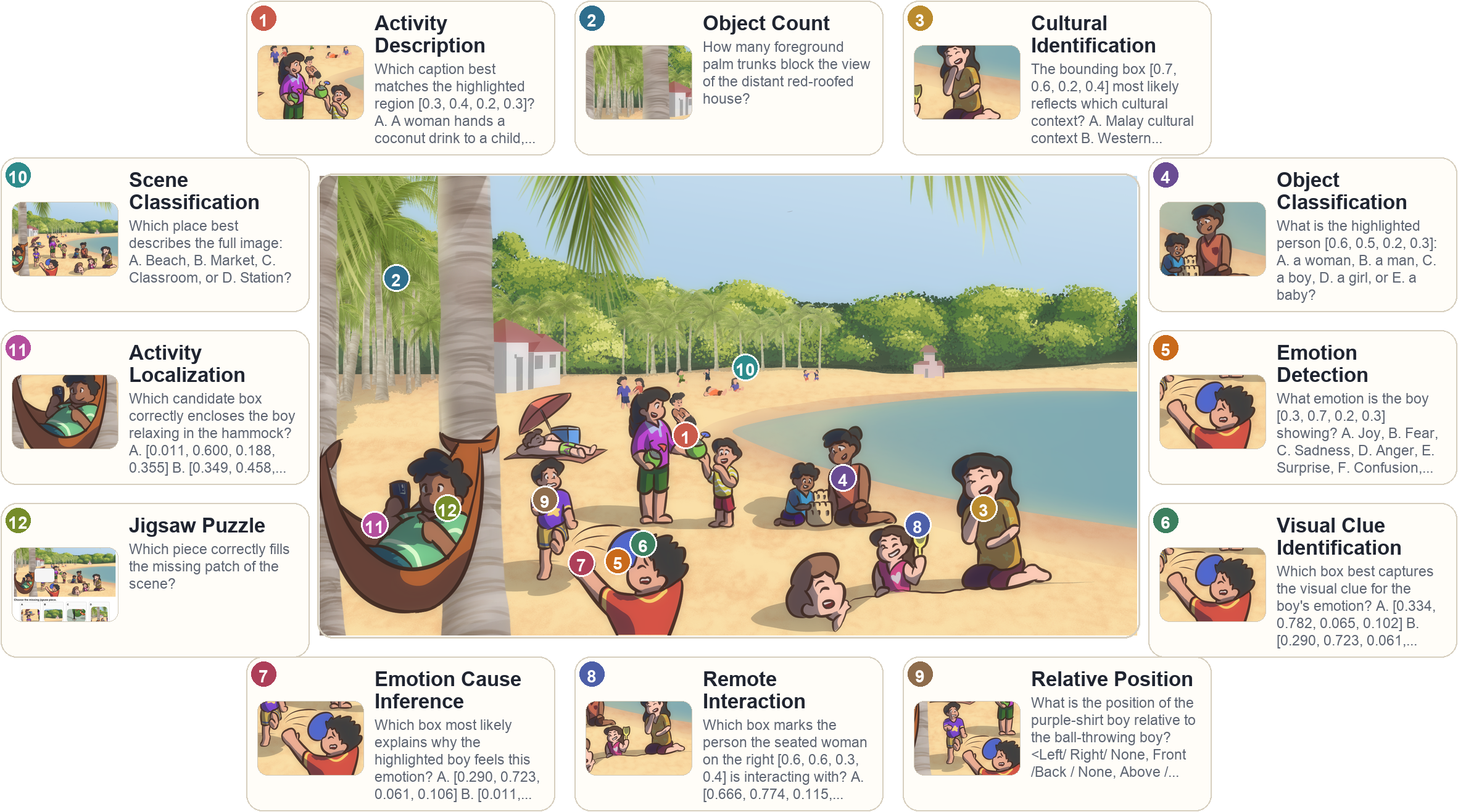}
    \caption{Overview of the 12 MUSE tasks. The cropped image on each task card is for illustration only. The model receives the full image and the corresponding question for all tasks except Jigsaw Puzzle, where it receives only the cropped image.}
    \label{fig:illustration}
\end{figure*}
Existing VLM benchmarks evaluate broad perception, knowledge, and reasoning abilities, including general multi-modal understanding~\cite{liu2024mmbench}, academic problem solving~\cite{lu2022scienceqa}, and scientific or mathematical reasoning~\cite{lu2024mathvista,ying2024mmtbench}. However, they are not designed to jointly assess the capabilities required for image-based language learning with artistic content.
Even when artistic content is included, the evaluation generally targets disciplinary knowledge or a specific aspect of art understanding~\cite{yue2024mmmu} rather than the capability required by educational VLMs. This leaves a gap between general VLM evaluation and the competencies needed to interact reliably with artistic educational imagery.

Benchmark construction also presents practical challenges. VLM benchmarks often construct task-specific question-answer pairs directly from individual images through manual annotation~\cite{zhang2025mme}. Extending such pipelines to new tasks requires additional annotation effort, while the resulting data are often difficult to reuse across tasks. Moreover, conventional question collection offers limited control over question form and complexity; prior work on controllable question generation shows that difficulty control requires explicit modeling of reasoning structure~\cite{cheng2021guiding}. Independently constructed tasks may also adopt inconsistent semantic representations, hindering comparability. We therefore decouple reusable visual-semantic annotations from task-specific question generation, improving scalability, consistency, and controllability.

To address both the evaluation and construction gaps, we introduce \textbf{MUSE}, a benchmark for \textbf{Multi-modal Understanding in Situated Education} using artistic imagery. MUSE adopts an \emph{annotation-first, task-generative} design: each artwork is annotated once with a reusable structured representation of its visual and semantic content, after which task-specific questions are instantiated through predefined generation rules. By separating \emph{what an image contains} from \emph{how a capability is queried}, this design supports annotation reuse, consistent semantics across tasks, and explicit control over question format and difficulty.

Built on this shared representation, MUSE turns each artwork into a multi-view evaluation instance. Its 12 tasks cover five complementary capability dimensions (Table~\ref{tab:capability_dimensions}) and combine textual and visual multiple-choice questions with numerical and open-ended responses. Tasks such as visual-clue identification and emotion-cause inference therefore test whether models can ground and articulate their understanding, rather than only recognize a correct option. Figure~\ref{fig:illustration} illustrates how one artwork supports the full task suite.

Evaluation of 30 open-source and proprietary VLMs reveals pronounced task-dependent gaps, particularly in visual grounding, affective interpretation, and compositional reasoning. Correlation and error analyses further show that success on general benchmarks or coarse recognition does not reliably transfer to artistic imagery and fine-grained evidence-based reasoning. Our main contributions are:

\begin{itemize}
\item We introduce MUSE, a 12-task benchmark that evaluates five dimensions of multimodal understanding over artistic imagery for image-based language learning and educational interaction.

\item We propose an annotation-first, task-generative construction framework that reuses structured image annotations to produce semantically consistent questions with controllable formats and difficulty.

\item We evaluate 30 open-source and proprietary VLMs on MUSE, revealing fundamental gaps between recognition, grounding, affective interpretation and compositional reasoning through task, correlation, and error analyses.
\end{itemize}

\section{Related Work}

\paragraph{Multimodal and educational benchmarks}
General VLM benchmarks evaluate perception, knowledge, and reasoning beyond conventional visual question answering. MMBench uses constructed multiple-choice questions (MCQs) for fine-grained assessment, while SEED-Bench uses human-verified questions to evaluate hierarchical capabilities~\cite{liu2024mmbench,li2024seedbench}. MMMU targets expert reasoning across disciplines; MMStar uses vision-indispensable samples to measure multimodal gain and leakage; and MMMU-Pro strengthens visual dependency through filtering, expanded options, and vision-only evaluation~\cite{yue2024mmmu,chen2024mmstar,yue2025mmmupro}. ScienceQA and MathVista focus on scientific and mathematical reasoning~\cite{lu2022scienceqa,lu2024mathvista}. These benchmarks primarily use natural images, diagrams, charts, documents, or examination materials, offering limited coverage of stylization, implicit narratives, affective evidence, and culturally situated meanings in artistic content for language learning.

\paragraph{Artistic, affective, and cultural understanding}
Prior work examines artistic, affective, and culturally grounded image understanding. ArtEmis collects emotion labels and visually grounded explanations for artworks, while ArtELingo adds multilingual annotations for cross-cultural affective responses~\cite{achlioptas2021artemis,mohamed2022artelingo}. VQArt-Bench evaluates symbolic meaning, narratives, counting, and visual relationships in art, whereas AICA-Bench addresses emotion understanding, reasoning, and generation~\cite{alfarano2025vqartbench,she2026aicabench}. CVQA evaluates culturally grounded visual question answering across regions and languages with native-speaker and expert data~\cite{romero2024cvqa}. These resources advance affective, artistic, or cultural understanding but generally focus on individual domains. MUSE instead jointly evaluates visual perception, activity and scene understanding, affective evidence and causes, spatial and compositional reasoning, and cultural understanding. Its decoupled construction reuses annotations across tasks, reduces annotation effort, and controls question formulation and difficulty.

\section{MUSE Benchmark}
\begin{figure}[!t]
    \centering
    \includegraphics[width=0.8\linewidth]{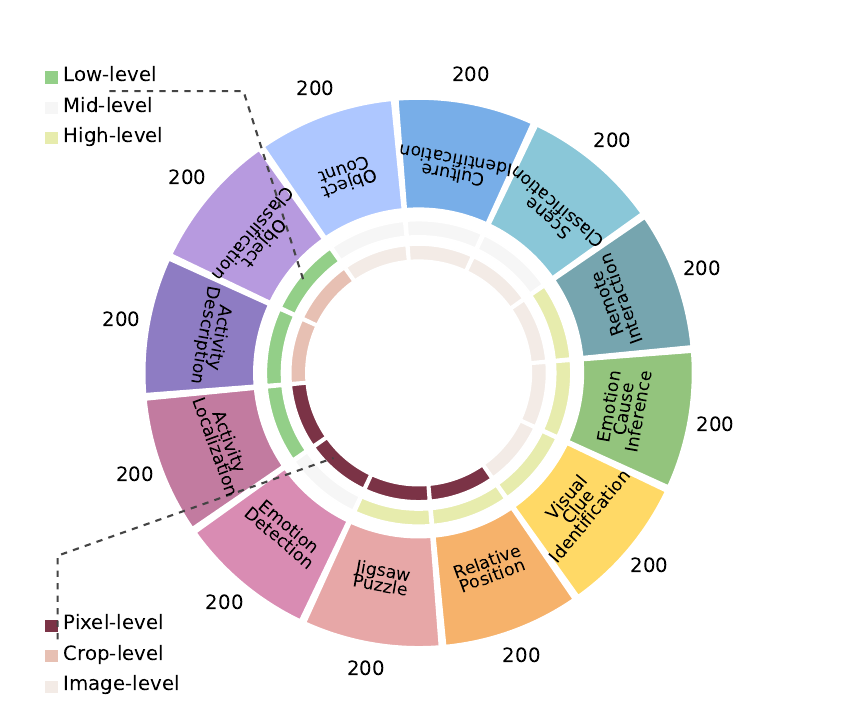}
    \caption{Taxonomy of MUSE and statistics.}
    \label{fig:taxonomy}
\end{figure}

MUSE differs from existing multimodal-understanding benchmarks in three ways: (1) it curates original artworks from artists worldwide to diversify image sources; (2) decouples annotation from question generation to control difficulty systematically; (3) and targets the visual capabilities required for reliable image-captioning-based language education. MUSE contains 2,400 questions over 1,174 images, each with a resolution of $1920 \times 1080$ pixels, across 12 tasks that test alignment between artistic visual content and linguistic descriptions. Figure~\ref{fig:taxonomy} shows the tasks span 3 cognitive complexity levels, i.e., low-level pattern recognition, mid-level semantic perception, and high-level reasoning, as well as 3 spatial granularities, i.e., pixel-, region-, and image-level understanding. Most use textual or visual multiple-choice questions; Object Count requires numerical prediction, while Visual Clue Identification and Emotion Cause Inference use open-ended responses evaluated by semantic similarity. We next describe its construction and tasks.

\subsection{Dataset Annotation and Quality Control}
Before annotation, 127 annotators receive a briefing on the study motivation, task definitions, guidelines, representative examples, and ambiguous cases. Using a standardized Label Studio Enterprise interface, they annotate activity, character, and object bounding boxes; emotion, object, and position labels; activity descriptions; scene and cultural labels; object counts; visual clues; and emotion causes. Each sample is independently annotated by one annotator, reviewed by two others, and finalized only after consensus, with disagreements resolved using the established guidelines.

\subsection{Question Generation}

\begin{figure}[!t]
    \centering
    \includegraphics[width=0.8\linewidth]{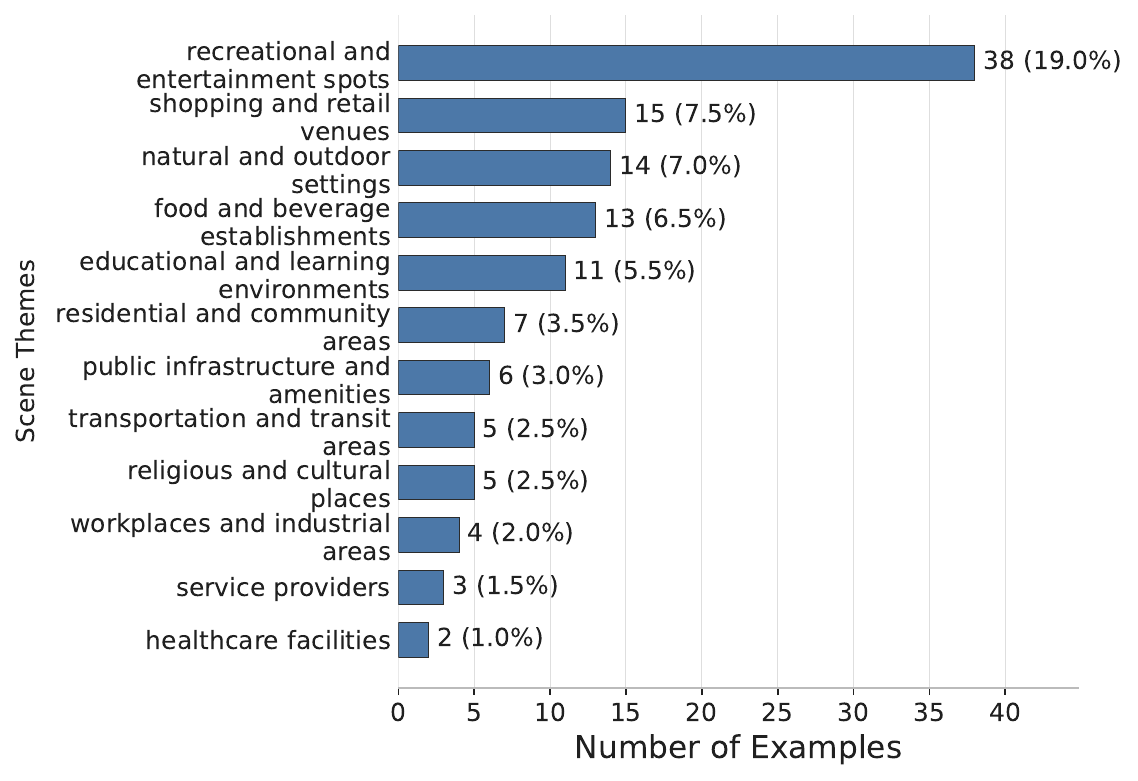}
    \caption{Scene theme distribution.}
    \label{fig:scene}
\end{figure}

To improve benchmark diversity, we explicitly enforce diversity along three dimensions during problem generation: artistic styles (through diverse artists), scene themes, and question difficulty. Scene theme distribution is in Figure~\ref{fig:scene}. Among these tasks, Object Classification, Emotion Detection, Visual Clue Identification, and Emotion Cause Inference form a four-turn sequence for evaluating affective computing, with questions and answers from earlier turns retained in the dialogue history. All bounding boxes below use normalized COCO format ($[x_{min}, y_{min}, width, height]$).

\begin{figure}[!t]
    \centering
    \includegraphics[width=0.8\linewidth]{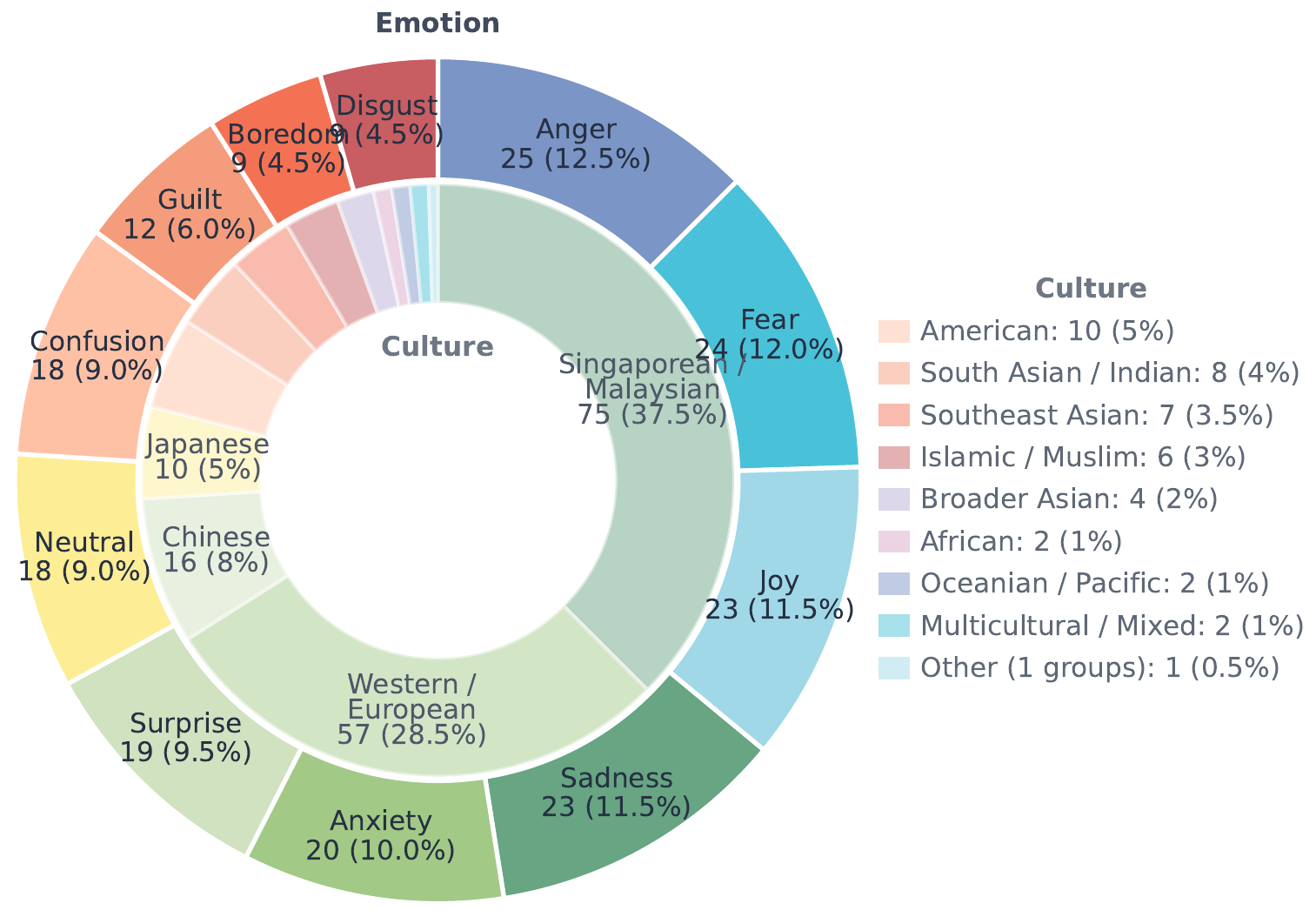}
    \caption{Emotion and culture distribution.}
    \label{fig:emotion and culture}
\end{figure}

\noindent\textsc{1. Object Classification}
Given a bounding box, models classify the character as Woman, Man, Girl, Boy, or Baby. The options are shuffled for each problem.

\noindent\textsc{2. Emotion Detection}
Models classify characters' emotion as Anxiety, Sadness, Surprise, Joy, Disgust, Fear, Boredom, Guilt, Neutral, Anger, or Confusion. The categories follow Plutchik's emotion wheel and primary, secondary, and tertiary dyads~\citep{plutchik1980general}, excluding emotions that are rare or difficult to depict visually. Options are shuffled, and the label distribution is in the outer ring of Figure~\ref{fig:emotion and culture}.

\noindent\textsc{3. Visual Clue Identification}
Models provide an open-ended description of the visual evidence supporting their preceding emotion prediction. Responses are compared with human references using semantic similarity.

\noindent\textsc{4. Emotion Cause Inference}
Models provide an open-ended explanation of the predicted emotion's cause, evaluated using the same metrics.

\noindent\textsc{5. Activity Localization}
Models select the bounding box corresponding to a described activity. Distractors comprise boxes for: i) another activity; ii) a character or inanimate object; iii) a subregion of the ground-truth box; iv) a random region; or v) ``None of the above.''

\noindent\textsc{6. Activity Description}
This task evaluate the VLMs' capability to understand and describe what is happening within the bounding boxes. 10 methods are employed to compose negative options: i) another activity description in the same image (oa); ii) another inanimate object in the same image (oosi); iii) another inanimate object in a different image (oodi); iv) another identity in the same image (oisi); v) another identity in a different image (oidi); vi) shifted the orders of objects in the original description (so); vii) concatenated $i$ activity descriptions in the same image ($i \in \{1,2,3\}$) (ca$\_$s); viii) concatenated $i$ activity descriptions in a different image ($i \in \{1,2,3\}$) (ca$\_$d); ix) negative descriptions from annotators (neg); and x) the statement "None of the above" (none).

\noindent\textsc{7. Cultural Identification}
Models identify cultural elements within a given bounding box. We embed all ground-truth labels using OpenAI \textsc{text-embedding-3-small} and cluster them into 15 categories. Three negative options are sampled from categories other than that of the ground truth. The inner ring of Figure~\ref{fig:emotion and culture} shows the category distribution.

\noindent\textsc{8. Jigsaw Puzzle}
Models complete jigsaw puzzles by aligning patches through continuity in shape, color, and texture. We use five segmentation grids: (3,4), (4,4), (3,6), (4,5), and (3,7). Distractors comprise: i) another piece from the same image; ii) the ground-truth piece combined with another piece; iii) a zoomed region around the ground-truth piece; or iv) a piece from another image. Pieces may be stretched, upright, or balanced hexagons; wide or landscape rectangles; thin-tall or portrait rectangles; or squares, with angled, rounded, or sharp edges.

\noindent\textsc{9. Object Count}
Models numerically predict object counts, testing object recognition and compositional reasoning under occlusion and variations in size and appearance.

\noindent\textsc{10. Relative Position}
Given object descriptions and bounding boxes, models predict three-dimensional spatial relations, particularly from the characters' viewpoints: i) left, none, or right laterally; ii) front, none, or back in depth; and iii) above, none, or under vertically.

\noindent\textsc{11. Remote Interaction}
Models reason about non-contact interactions between entities localized by descriptions and bounding boxes. Each query contains two MCQs: one identifies the interacting entity, and the other identifies supporting visual evidence. Distractors comprise: i) entities from other interactions in the same image; ii) evidence from other same-image interactions; iii) mismatched text--bounding-box pairs sampled from these candidates and the ground truth; and iv) cross-image candidates with different descriptions and low overlap with the ground-truth box.

\noindent\textsc{12. Scene Classification} 
Models classify the overall scene by integrating global visual and semantic information. We use OpenAI \textsc{gpt-3.5} to organize all ground-truth scene labels into 13 categories, then generate three negative options by sampling one label from each of three categories other than the ground-truth category.
\section{Experiments}
\begin{table*}[!h]
\centering

\scriptsize

\resizebox{\textwidth}{!}{
\begin{tabular}{@{}lcccccccccccc@{}}
\toprule

& \multicolumn{2}{c}{\textbf{Visual Perception}}
& \multicolumn{3}{c}{\textbf{Semantic Understanding}}
& \multicolumn{3}{c}{\textbf{Affective Interpretation}}
& \multicolumn{3}{c}{\textbf{Compositional Reasoning}}
& \multicolumn{1}{c}{\textbf{Cultural}} \\

\cmidrule(lr){2-3}
\cmidrule(lr){4-6}
\cmidrule(lr){7-9}
\cmidrule(lr){10-12}
\cmidrule(lr){13-13}

\textbf{Model}
& \makecell{\textbf{Object}\\\textbf{Cls.}}
& \makecell{\textbf{Object}\\\textbf{Count}}
& \makecell{\textbf{Activity}\\\textbf{Loc.}}
& \makecell{\textbf{Activity}\\\textbf{Desc.}}
& \makecell{\textbf{Scene}\\\textbf{Cls.}}
& \makecell{\textbf{Emotion}\\\textbf{Det.}}
& \makecell{\textbf{Visual Clue}\\\textbf{Ident.}}
& \makecell{\textbf{Emotion Cause}\\\textbf{Infer.}}
& \makecell{\textbf{Relative}\\\textbf{Position}}
& \makecell{\textbf{Remote}\\\textbf{Interaction}}
& \makecell{\textbf{Jigsaw}\\\textbf{Puzzle}}
& \makecell{\textbf{Cultural}\\\textbf{Ident.}} \\
\midrule

GPT-5.6-Sol
& \cellcolor{MUEHeatFive}\textbf{76.0}
& \cellcolor{MUEHeatFive}\textbf{71.5}
& \cellcolor{MUEHeatFour}\underline{69.5}
& \cellcolor{MUEHeatTwo}34.5
& \cellcolor{MUEHeatFive}\underline{86.5}
& \cellcolor{MUEHeatThree}\textbf{39.5}
& \cellcolor{MUEHeatThree}\textbf{50.90}
& \cellcolor{MUEHeatThree}\textbf{49.18}
& \cellcolor{MUEHeatOne}4.0
& \cellcolor{MUEHeatFive}\textbf{86.5}
& \cellcolor{MUEHeatThree}\underline{35.5}
& \cellcolor{MUEHeatFive}\textbf{76.5} \\
Qwen3-VL-32b
& \cellcolor{MUEHeatFour}\underline{54.0}
& \cellcolor{MUEHeatFour}\underline{59.0}
& \cellcolor{MUEHeatFive}\textbf{72.5}
& \cellcolor{MUEHeatThree}\underline{50.0}
& \cellcolor{MUEHeatFive}\textbf{87.0}
& \cellcolor{MUEHeatTwo}\underline{29.5}
& \cellcolor{MUEHeatThree}\underline{44.34}
& \cellcolor{MUEHeatThree}\underline{40.23}
& \cellcolor{MUEHeatOne}5.0
& \cellcolor{MUEHeatThree}\underline{52.0}
& \cellcolor{MUEHeatTwo}28.5
& \cellcolor{MUEHeatFour}\underline{60.0} \\
InternVL3-38b
& \cellcolor{MUEHeatThree}49.0
& \cellcolor{MUEHeatThree}47.0
& \cellcolor{MUEHeatFour}61.5
& \cellcolor{MUEHeatThree}38.5
& \cellcolor{MUEHeatFive}85.0
& \cellcolor{MUEHeatTwo}18.0
& \cellcolor{MUEHeatThree}35.24
& \cellcolor{MUEHeatTwo}30.95
& \cellcolor{MUEHeatOne}\underline{8.5}
& \cellcolor{MUEHeatThree}43.0
& \cellcolor{MUEHeatTwo}31.5
& \cellcolor{MUEHeatThree}51.0 \\
Qwen2.5-VL-72b
& \cellcolor{MUEHeatThree}52.0
& \cellcolor{MUEHeatThree}51.0
& \cellcolor{MUEHeatFour}56.0
& \cellcolor{MUEHeatThree}47.0
& \cellcolor{MUEHeatFive}86.0
& \cellcolor{MUEHeatTwo}23.5
& \cellcolor{MUEHeatThree}40.26
& \cellcolor{MUEHeatTwo}33.63
& \cellcolor{MUEHeatOne}\underline{8.5}
& \cellcolor{MUEHeatThree}46.0
& \cellcolor{MUEHeatTwo}21.5
& \cellcolor{MUEHeatThree}50.5 \\
Qwen3-VL-8b
& \cellcolor{MUEHeatThree}46.5
& \cellcolor{MUEHeatThree}48.5
& \cellcolor{MUEHeatFour}58.0
& \cellcolor{MUEHeatThree}42.5
& \cellcolor{MUEHeatFive}84.5
& \cellcolor{MUEHeatTwo}25.0
& \cellcolor{MUEHeatThree}44.06
& \cellcolor{MUEHeatThree}38.47
& \cellcolor{MUEHeatOne}2.0
& \cellcolor{MUEHeatThree}40.5
& \cellcolor{MUEHeatOne}16.5
& \cellcolor{MUEHeatFour}58.5 \\
InternVL3-14b
& \cellcolor{MUEHeatThree}48.0
& \cellcolor{MUEHeatThree}44.5
& \cellcolor{MUEHeatFour}55.0
& \cellcolor{MUEHeatThree}44.5
& \cellcolor{MUEHeatFive}81.5
& \cellcolor{MUEHeatOne}15.0
& \cellcolor{MUEHeatTwo}34.25
& \cellcolor{MUEHeatTwo}27.59
& \cellcolor{MUEHeatOne}\textbf{10.0}
& \cellcolor{MUEHeatThree}39.0
& \cellcolor{MUEHeatTwo}28.0
& \cellcolor{MUEHeatThree}46.5 \\
GPT-4o
& \cellcolor{MUEHeatTwo}24.5
& \cellcolor{MUEHeatThree}49.0
& \cellcolor{MUEHeatThree}51.5
& \cellcolor{MUEHeatFour}\textbf{59.5}
& \cellcolor{MUEHeatFive}86.0
& \cellcolor{MUEHeatTwo}22.0
& \cellcolor{MUEHeatThree}34.84
& \cellcolor{MUEHeatTwo}27.97
& \cellcolor{MUEHeatOne}7.5
& \cellcolor{MUEHeatTwo}30.0
& \cellcolor{MUEHeatTwo}28.0
& \cellcolor{MUEHeatThree}39.5 \\
Qwen2.5-VL-32b
& \cellcolor{MUEHeatThree}44.5
& \cellcolor{MUEHeatThree}50.0
& \cellcolor{MUEHeatThree}45.0
& \cellcolor{MUEHeatTwo}34.5
& \cellcolor{MUEHeatFive}83.0
& \cellcolor{MUEHeatTwo}21.5
& \cellcolor{MUEHeatThree}39.65
& \cellcolor{MUEHeatTwo}34.27
& \cellcolor{MUEHeatOne}7.5
& \cellcolor{MUEHeatThree}44.0
& \cellcolor{MUEHeatTwo}20.5
& \cellcolor{MUEHeatThree}51.5 \\
Qwen2.5-VL-7b
& \cellcolor{MUEHeatTwo}26.0
& \cellcolor{MUEHeatThree}42.0
& \cellcolor{MUEHeatThree}50.0
& \cellcolor{MUEHeatThree}36.5
& \cellcolor{MUEHeatFive}83.0
& \cellcolor{MUEHeatOne}15.5
& \cellcolor{MUEHeatThree}37.99
& \cellcolor{MUEHeatTwo}28.77
& \cellcolor{MUEHeatOne}2.5
& \cellcolor{MUEHeatTwo}32.0
& \cellcolor{MUEHeatTwo}19.0
& \cellcolor{MUEHeatFour}54.5 \\
Gemma3-12b-it
& \cellcolor{MUEHeatTwo}24.0
& \cellcolor{MUEHeatThree}40.0
& \cellcolor{MUEHeatThree}39.5
& \cellcolor{MUEHeatThree}35.0
& \cellcolor{MUEHeatFive}81.0
& \cellcolor{MUEHeatTwo}17.5
& \cellcolor{MUEHeatThree}41.58
& \cellcolor{MUEHeatTwo}33.66
& \cellcolor{MUEHeatOne}1.5
& \cellcolor{MUEHeatTwo}26.5
& \cellcolor{MUEHeatTwo}24.5
& \cellcolor{MUEHeatFour}55.0 \\
Gemma3-27b-it
& \cellcolor{MUEHeatTwo}32.0
& \cellcolor{MUEHeatThree}42.0
& \cellcolor{MUEHeatThree}47.0
& \cellcolor{MUEHeatTwo}30.5
& \cellcolor{MUEHeatFive}81.0
& \cellcolor{MUEHeatTwo}18.0
& \cellcolor{MUEHeatThree}42.21
& \cellcolor{MUEHeatThree}35.74
& \cellcolor{MUEHeatOne}4.5
& \cellcolor{MUEHeatTwo}24.5
& \cellcolor{MUEHeatTwo}19.5
& \cellcolor{MUEHeatThree}48.0 \\
InternVL3-9b
& \cellcolor{MUEHeatTwo}20.0
& \cellcolor{MUEHeatThree}44.0
& \cellcolor{MUEHeatThree}46.0
& \cellcolor{MUEHeatThree}35.5
& \cellcolor{MUEHeatFive}81.5
& \cellcolor{MUEHeatOne}9.0
& \cellcolor{MUEHeatThree}37.56
& \cellcolor{MUEHeatTwo}28.59
& \cellcolor{MUEHeatOne}2.5
& \cellcolor{MUEHeatTwo}23.0
& \cellcolor{MUEHeatTwo}31.5
& \cellcolor{MUEHeatThree}47.5 \\
MiniCPM-V-2.6
& \cellcolor{MUEHeatTwo}30.5
& \cellcolor{MUEHeatThree}46.0
& \cellcolor{MUEHeatThree}43.0
& \cellcolor{MUEHeatThree}41.0
& \cellcolor{MUEHeatFive}84.5
& \cellcolor{MUEHeatOne}16.0
& \cellcolor{MUEHeatThree}40.08
& \cellcolor{MUEHeatTwo}28.79
& \cellcolor{MUEHeatOne}2.5
& \cellcolor{MUEHeatOne}9.0
& \cellcolor{MUEHeatOne}12.0
& \cellcolor{MUEHeatThree}43.0 \\
DeepSeek-VL2
& \cellcolor{MUEHeatTwo}29.5
& \cellcolor{MUEHeatThree}50.0
& \cellcolor{MUEHeatTwo}33.5
& \cellcolor{MUEHeatTwo}22.5
& \cellcolor{MUEHeatFive}83.0
& \cellcolor{MUEHeatOne}17.0
& \cellcolor{MUEHeatThree}40.15
& \cellcolor{MUEHeatTwo}32.10
& \cellcolor{MUEHeatOne}3.0
& \cellcolor{MUEHeatOne}12.0
& \cellcolor{MUEHeatTwo}25.0
& \cellcolor{MUEHeatThree}50.0 \\
InternVL3-8b
& \cellcolor{MUEHeatTwo}33.0
& \cellcolor{MUEHeatThree}41.0
& \cellcolor{MUEHeatThree}44.0
& \cellcolor{MUEHeatTwo}21.0
& \cellcolor{MUEHeatFive}82.5
& \cellcolor{MUEHeatOne}13.0
& \cellcolor{MUEHeatThree}34.93
& \cellcolor{MUEHeatTwo}27.98
& \cellcolor{MUEHeatOne}1.0
& \cellcolor{MUEHeatTwo}23.5
& \cellcolor{MUEHeatTwo}17.5
& \cellcolor{MUEHeatThree}49.0 \\
MiniCPM-Llama3-V-2.5
& \cellcolor{MUEHeatTwo}28.5
& \cellcolor{MUEHeatThree}35.0
& \cellcolor{MUEHeatThree}42.5
& \cellcolor{MUEHeatThree}36.5
& \cellcolor{MUEHeatFive}75.5
& \cellcolor{MUEHeatOne}11.0
& \cellcolor{MUEHeatThree}36.53
& \cellcolor{MUEHeatTwo}24.67
& \cellcolor{MUEHeatOne}3.0
& \cellcolor{MUEHeatOne}16.5
& \cellcolor{MUEHeatTwo}28.5
& \cellcolor{MUEHeatThree}47.5 \\
MiniCPM-O-2.6
& \cellcolor{MUEHeatTwo}34.5
& \cellcolor{MUEHeatThree}47.0
& \cellcolor{MUEHeatThree}47.5
& \cellcolor{MUEHeatTwo}26.0
& \cellcolor{MUEHeatFive}81.0
& \cellcolor{MUEHeatTwo}19.5
& \cellcolor{MUEHeatThree}36.26
& \cellcolor{MUEHeatTwo}27.09
& \cellcolor{MUEHeatOne}3.0
& \cellcolor{MUEHeatOne}8.5
& \cellcolor{MUEHeatOne}17.0
& \cellcolor{MUEHeatThree}47.0 \\
GLM-4V-9b
& \cellcolor{MUEHeatTwo}26.0
& \cellcolor{MUEHeatTwo}29.0
& \cellcolor{MUEHeatThree}45.0
& \cellcolor{MUEHeatTwo}25.0
& \cellcolor{MUEHeatFour}64.5
& \cellcolor{MUEHeatOne}14.0
& \cellcolor{MUEHeatThree}37.07
& \cellcolor{MUEHeatTwo}24.60
& \cellcolor{MUEHeatOne}2.0
& \cellcolor{MUEHeatTwo}27.5
& \cellcolor{MUEHeatThree}\textbf{42.0}
& \cellcolor{MUEHeatThree}51.5 \\
Qwen2.5-VL-3b
& \cellcolor{MUEHeatOne}13.0
& \cellcolor{MUEHeatThree}45.0
& \cellcolor{MUEHeatThree}36.0
& \cellcolor{MUEHeatThree}37.0
& \cellcolor{MUEHeatFive}77.5
& \cellcolor{MUEHeatOne}12.5
& \cellcolor{MUEHeatTwo}33.50
& \cellcolor{MUEHeatTwo}20.78
& \cellcolor{MUEHeatOne}6.0
& \cellcolor{MUEHeatOne}9.5
& \cellcolor{MUEHeatOne}14.5
& \cellcolor{MUEHeatThree}45.0 \\
LLaVA-Next-8b
& \cellcolor{MUEHeatThree}38.0
& \cellcolor{MUEHeatTwo}31.5
& \cellcolor{MUEHeatThree}44.5
& \cellcolor{MUEHeatTwo}22.5
& \cellcolor{MUEHeatFour}61.0
& \cellcolor{MUEHeatOne}12.0
& \cellcolor{MUEHeatTwo}33.34
& \cellcolor{MUEHeatTwo}27.63
& \cellcolor{MUEHeatOne}1.5
& \cellcolor{MUEHeatOne}8.5
& \cellcolor{MUEHeatTwo}21.5
& \cellcolor{MUEHeatThree}41.5 \\
DeepSeek-VL2-Small
& \cellcolor{MUEHeatTwo}18.0
& \cellcolor{MUEHeatThree}38.0
& \cellcolor{MUEHeatTwo}31.0
& \cellcolor{MUEHeatTwo}23.0
& \cellcolor{MUEHeatFive}75.5
& \cellcolor{MUEHeatOne}15.5
& \cellcolor{MUEHeatThree}41.05
& \cellcolor{MUEHeatTwo}31.63
& \cellcolor{MUEHeatOne}2.0
& \cellcolor{MUEHeatOne}12.0
& \cellcolor{MUEHeatTwo}21.5
& \cellcolor{MUEHeatThree}44.0 \\
Gemma3-4b-it
& \cellcolor{MUEHeatTwo}24.5
& \cellcolor{MUEHeatTwo}28.0
& \cellcolor{MUEHeatTwo}32.0
& \cellcolor{MUEHeatTwo}21.0
& \cellcolor{MUEHeatFive}81.0
& \cellcolor{MUEHeatOne}14.5
& \cellcolor{MUEHeatThree}39.02
& \cellcolor{MUEHeatTwo}29.36
& \cellcolor{MUEHeatOne}2.0
& \cellcolor{MUEHeatOne}11.0
& \cellcolor{MUEHeatTwo}21.0
& \cellcolor{MUEHeatThree}43.5 \\
InternVL3-2b
& \cellcolor{MUEHeatTwo}32.5
& \cellcolor{MUEHeatTwo}32.5
& \cellcolor{MUEHeatTwo}26.0
& \cellcolor{MUEHeatTwo}23.5
& \cellcolor{MUEHeatFive}78.5
& \cellcolor{MUEHeatOne}11.0
& \cellcolor{MUEHeatTwo}34.15
& \cellcolor{MUEHeatTwo}27.66
& \cellcolor{MUEHeatOne}3.0
& \cellcolor{MUEHeatOne}9.0
& \cellcolor{MUEHeatTwo}24.5
& \cellcolor{MUEHeatThree}37.0 \\
Yi-VL-6b
& \cellcolor{MUEHeatTwo}18.5
& \cellcolor{MUEHeatOne}16.5
& \cellcolor{MUEHeatThree}35.5
& \cellcolor{MUEHeatThree}45.0
& \cellcolor{MUEHeatFive}77.0
& \cellcolor{MUEHeatOne}11.5
& \cellcolor{MUEHeatTwo}25.51
& \cellcolor{MUEHeatTwo}24.98
& \cellcolor{MUEHeatOne}0.5
& \cellcolor{MUEHeatOne}3.5
& \cellcolor{MUEHeatTwo}21.5
& \cellcolor{MUEHeatTwo}34.5 \\
InternVL3-1b
& \cellcolor{MUEHeatTwo}27.5
& \cellcolor{MUEHeatTwo}33.5
& \cellcolor{MUEHeatThree}35.5
& \cellcolor{MUEHeatTwo}19.5
& \cellcolor{MUEHeatFive}71.5
& \cellcolor{MUEHeatOne}11.5
& \cellcolor{MUEHeatTwo}32.16
& \cellcolor{MUEHeatTwo}21.35
& \cellcolor{MUEHeatOne}7.5
& \cellcolor{MUEHeatOne}7.0
& \cellcolor{MUEHeatTwo}25.5
& \cellcolor{MUEHeatTwo}31.0 \\
Yi-VL-34b
& \cellcolor{MUEHeatTwo}27.0
& \cellcolor{MUEHeatTwo}24.5
& \cellcolor{MUEHeatTwo}23.5
& \cellcolor{MUEHeatTwo}28.5
& \cellcolor{MUEHeatFour}63.0
& \cellcolor{MUEHeatOne}9.0
& \cellcolor{MUEHeatTwo}33.31
& \cellcolor{MUEHeatTwo}25.00
& \cellcolor{MUEHeatOne}0.0
& \cellcolor{MUEHeatTwo}20.0
& \cellcolor{MUEHeatTwo}22.0
& \cellcolor{MUEHeatTwo}29.5 \\
CogVLM2-19b
& \cellcolor{MUEHeatTwo}19.5
& \cellcolor{MUEHeatTwo}26.0
& \cellcolor{MUEHeatTwo}18.5
& \cellcolor{MUEHeatOne}12.0
& \cellcolor{MUEHeatFive}\underline{86.5}
& \cellcolor{MUEHeatOne}12.5
& \cellcolor{MUEHeatThree}37.23
& \cellcolor{MUEHeatTwo}25.12
& \cellcolor{MUEHeatOne}3.0
& \cellcolor{MUEHeatOne}8.5
& \cellcolor{MUEHeatTwo}21.5
& \cellcolor{MUEHeatThree}39.5 \\
LLaVA-Next-34b
& \cellcolor{MUEHeatTwo}23.5
& \cellcolor{MUEHeatOne}0.0
& \cellcolor{MUEHeatThree}40.0
& \cellcolor{MUEHeatThree}36.5
& \cellcolor{MUEHeatFour}53.5
& \cellcolor{MUEHeatOne}9.0
& \cellcolor{MUEHeatThree}36.06
& \cellcolor{MUEHeatTwo}17.67
& \cellcolor{MUEHeatOne}2.5
& \cellcolor{MUEHeatOne}12.0
& \cellcolor{MUEHeatTwo}21.5
& \cellcolor{MUEHeatTwo}19.0 \\
LLaVA-Next-72b
& \cellcolor{MUEHeatTwo}23.0
& \cellcolor{MUEHeatTwo}29.5
& \cellcolor{MUEHeatFour}54.5
& \cellcolor{MUEHeatTwo}24.5
& \cellcolor{MUEHeatTwo}30.5
& \cellcolor{MUEHeatOne}9.0
& \cellcolor{MUEHeatTwo}20.87
& \cellcolor{MUEHeatOne}9.31
& \cellcolor{MUEHeatOne}3.0
& \cellcolor{MUEHeatOne}17.0
& \cellcolor{MUEHeatTwo}27.0
& \cellcolor{MUEHeatTwo}19.0 \\
DeepSeek-VL2-Tiny
& \cellcolor{MUEHeatTwo}17.5
& \cellcolor{MUEHeatTwo}31.0
& \cellcolor{MUEHeatOne}14.5
& \cellcolor{MUEHeatTwo}25.5
& \cellcolor{MUEHeatFour}53.0
& \cellcolor{MUEHeatOne}14.0
& \cellcolor{MUEHeatThree}37.11
& \cellcolor{MUEHeatTwo}17.59
& \cellcolor{MUEHeatOne}0.0
& \cellcolor{MUEHeatOne}1.0
& \cellcolor{MUEHeatTwo}21.5
& \cellcolor{MUEHeatTwo}34.0 \\
\bottomrule
\end{tabular}
}

\begin{minipage}{0.99\textwidth}
\scriptsize
\textit{Note:}
Object Cls.: Object Classification;
Activity Loc.: Activity Localization;
Activity Desc.: Activity Description;
Scene Cls.: Scene Classification;
Emotion Det.: Emotion Detection;
Visual Clue Ident.: Visual Clue Identification;
Emotion Cause Infer.: Emotion Cause Inference.
Visual Clue Identification and Emotion Cause Inference are evaluated using
semantic similarity scores.
\end{minipage}
\caption{Performance on the 12 MUSE tasks, with models ordered by average performance.
All results are reported as percentages.
The best and second-best results in each column are highlighted in
\textbf{bold} and \underline{underlined}, respectively.}
\label{tab:main-results}
\end{table*}

\begin{figure}[!h]
    \centering
    \includegraphics[width=0.8\linewidth]{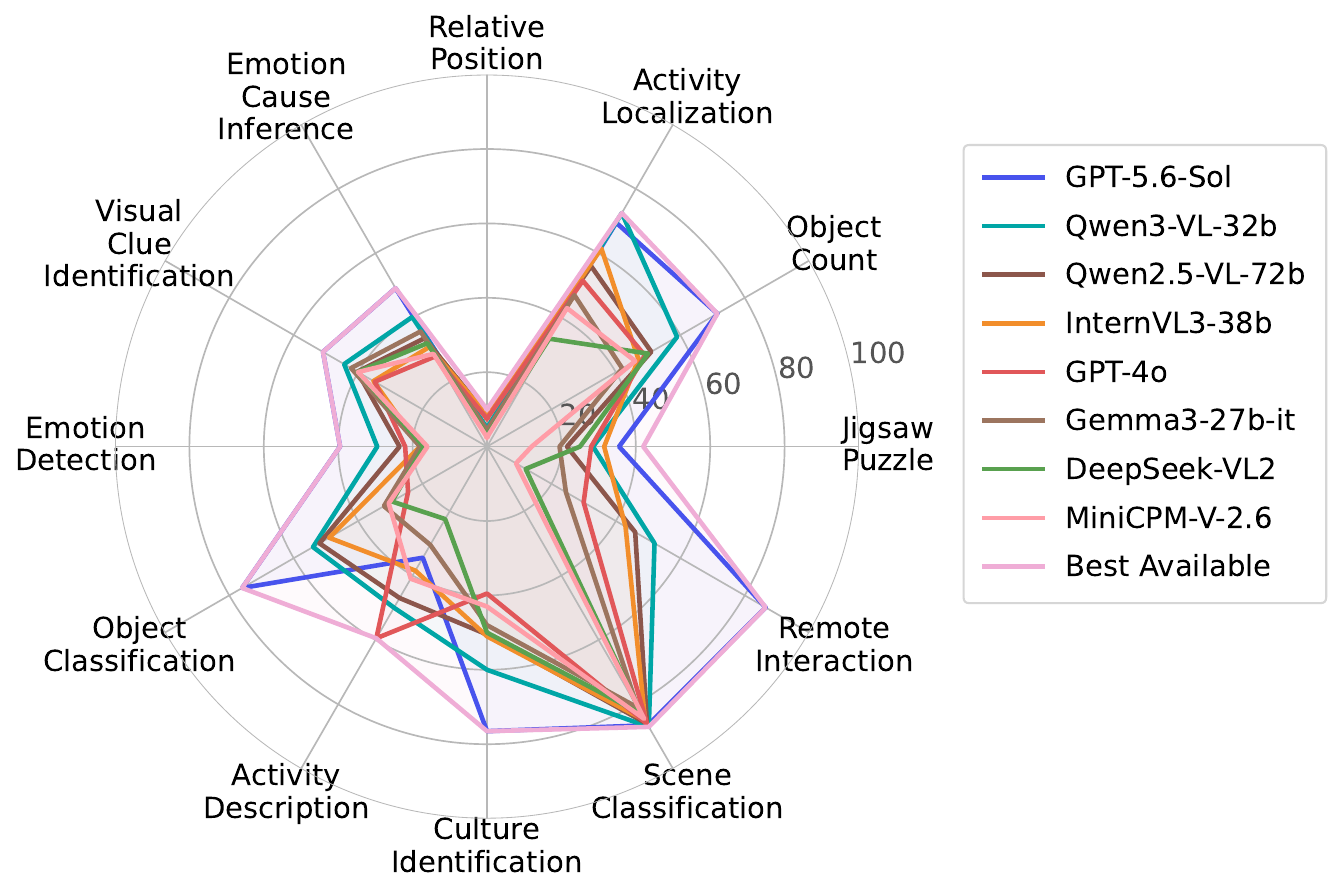}
    \caption{Top representatives from 8 model families; ``Best Available'' shows the per-task maximum across models.}
    \label{fig:best models}
\end{figure}
We evaluate 30 open-source and proprietary multimodal models spanning architectures, scales, and training paradigms. GPT-5.6-Sol and GPT-4o are accessed through APIs, while open-source models are deployed on AWS instances equipped with NVIDIA T4, A10G, or A100 GPUs. The evaluated families include CogVLM2~\cite{hong2024cogvlm2}, DeepSeek-VL2~\cite{wu2024deepseekvl2}, Gemma 3~\cite{gemmateam2025gemma3}, GLM-4V~\cite{hong2024cogvlm2}, InternVL3~\cite{zhu2025internvl3}, LLaVA-NeXT~\cite{liu2024llavanext}, MiniCPM-V~\cite{yao2024minicpmv}, MiniCPM-o~\cite{openbmb2025minicpmo}, Qwen2.5-VL~\cite{bai2025qwen25vl}, Qwen3-VL~\cite{bai2025qwen3vl}, and Yi-VL~\cite{young2024yi}. All models use temperature 0 and are evaluated once as their outputs are stable. A unified parser handles free-form, option-based, and JSON responses; tasks are scored by accuracy or semantic similarity (i.e., cosine similarity between \textsc{text-embedding-3-large} embeddings).

\subsection{Main Results}
Table~\ref{tab:main-results} shows that \textit{performance remains highly task-dependent, with no model dominating across all capabilities}. GPT-5.6-Sol achieves the strongest overall results and surpasses GPT-4o on 10 of 12 tasks, yet GPT-4o remains superior on Activity Description and Relative Position. Open-source models also retain task-specific advantages: Qwen3-VL-32b leads Activity Localization and Scene Classification, while InternVL3-14b and GLM-4V-9b perform best on Relative Position and Jigsaw Puzzle, respectively. These results indicate that \textit{progress is uneven and does not translate uniformly across capability dimensions}.

A clear divide emerges between \textit{recognition and integrative reasoning}. Scene Classification is comparatively mature, with 23 of 30 models exceeding 75.0 and a median score of 81.0. In contrast, Emotion Detection, Relative Position, Remote Interaction, and Jigsaw Puzzle exhibit substantially lower medians, revealing \textit{persistent limitations in affective interpretation, and compositional reasoning}.

Figure~\ref{fig:best models} further shows that model families share similar strengths in scene and activity recognition but diverge sharply on Activity Description, Remote Interaction, and Jigsaw Puzzle, suggesting that \textit{architecture and training remain important determinants of capability-specific performance}. Scaling is also \textit{non-monotonic}: although InternVL3-38b outperforms InternVL3-14b on most tasks, it performs worse on Activity Description and Relative Position. Overall, \textit{current VLMs are more reliable at recognizing visible content than at grounding predictions in visual evidence, explaining affective causes, or reasoning over perspective-dependent and non-contact relations}.

\subsection{Inter-Task Correlation Analysis}
\begin{figure}[t]
    \centering
    \includegraphics[width=0.8\linewidth]{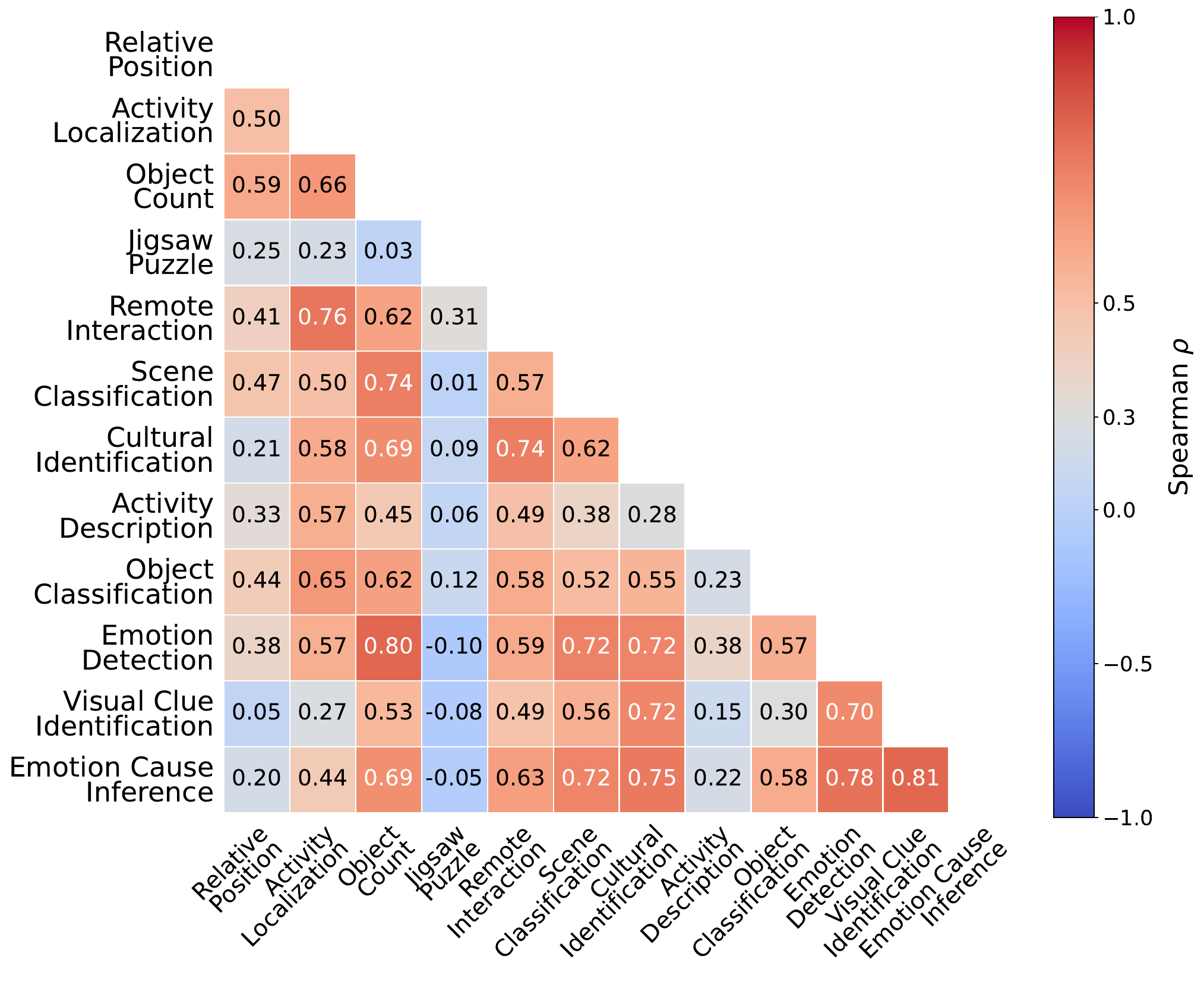}
    \caption{Spearman rank correlations among 12 MUSE tasks.}
    \label{fig:task-correlations}
\end{figure}
We compute pairwise Spearman's $\rho$ across 30 models to examine relationships among tasks. Figure~\ref{fig:task-correlations} reveals several \textit{coherent capability} groups: Object Count, Emotion Detection, and Scene Classification are strongly correlated, while Visual Clue Identification closely tracks Emotion Cause Inference, linking visual evidence grounding with affective reasoning. Activity Localization, Activity Description, and Remote Interaction form a moderately correlated group centered on entity-activity and cross-region reasoning. In contrast, Jigsaw Puzzle correlates weakly with most tasks, indicating a distinct compositional capability. Overall, MUSE captures \textit{related but non-redundant dimensions} of multimodal understanding rather than a single underlying competence.
\begin{figure}[h]
    \centering
    \includegraphics[width=0.8\linewidth]{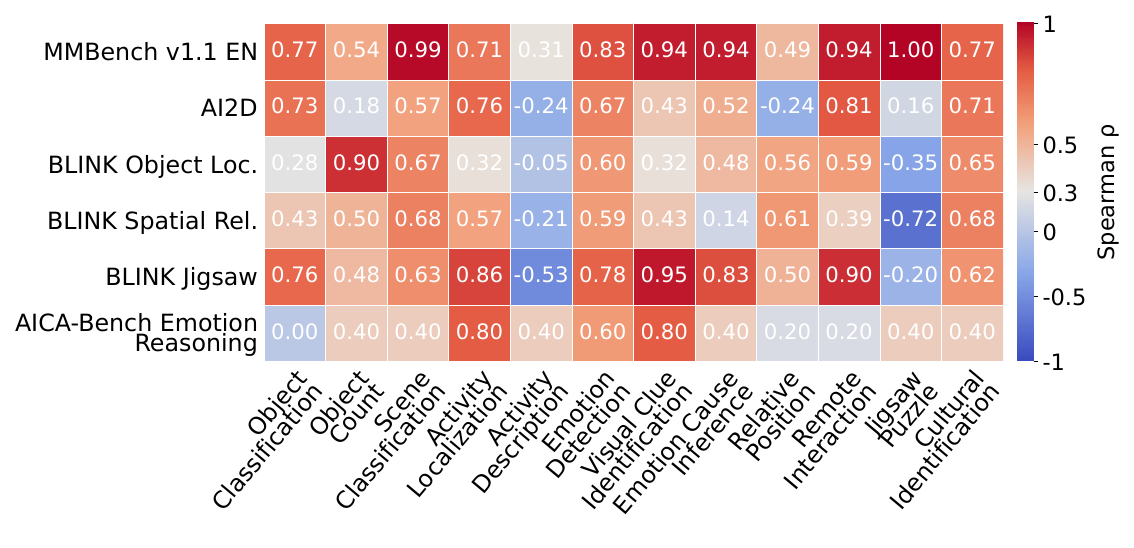}
    \caption{Spearman rank correlations between 6 existing benchmarks and 12 MUSE tasks.}
    \label{fig:external-correlations}
\end{figure}
\subsection{Complementarity to Existing Benchmarks}
Using pairwise-available model scores, we compute Spearman's $\rho$ between the 12 MUSE tasks and 6 external benchmarks.

\subsection{Cross-Task Error Analysis}
\begin{figure}[!h]
    \centering
    \begin{minipage}[!h]{0.5\textwidth}
        \centering
        \includegraphics[width=\linewidth]{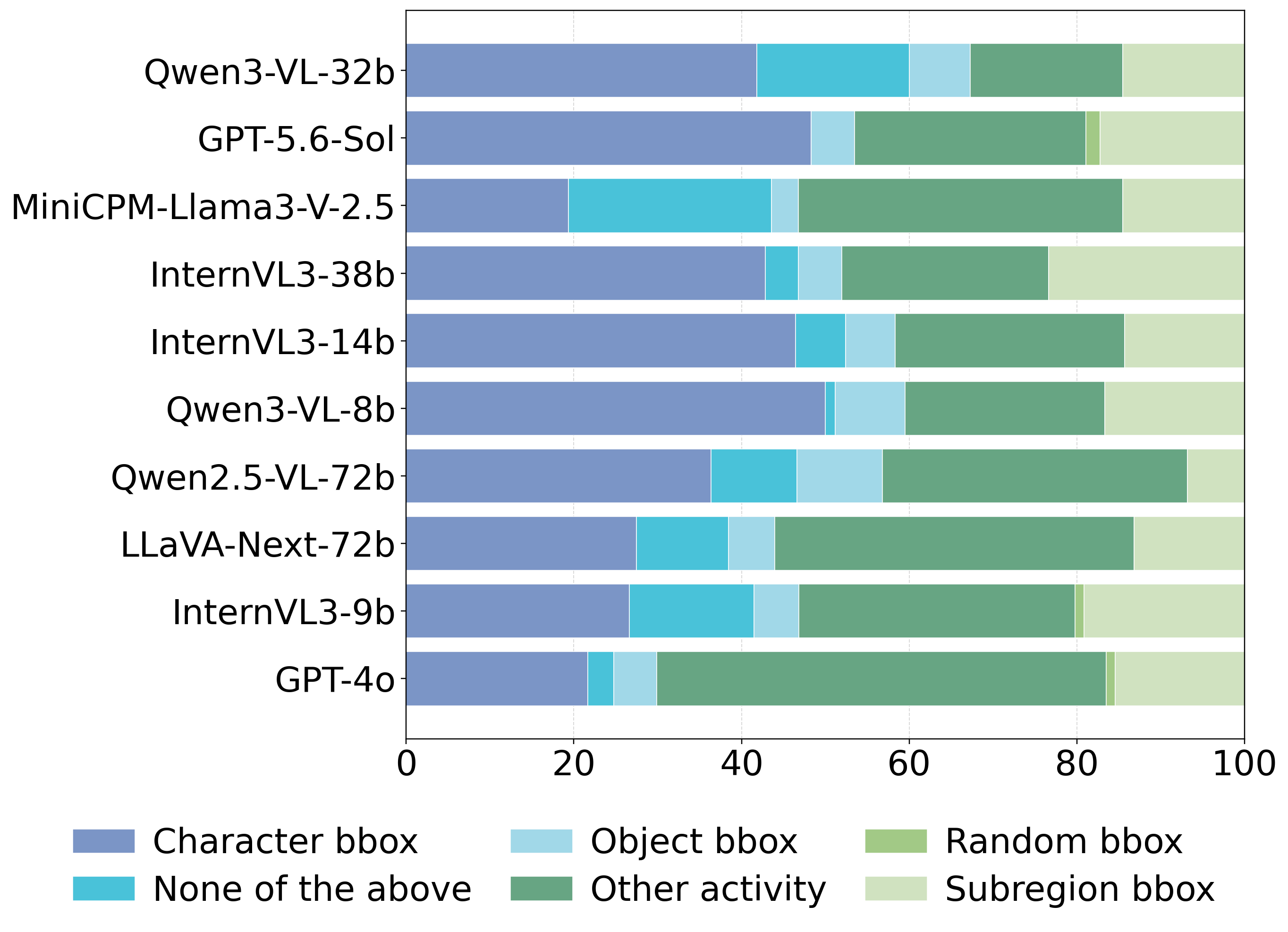}
        \footnotesize (a) Activity Localization
    \end{minipage}\hfill
    \begin{minipage}[!h]{0.5\textwidth}
        \centering
        \includegraphics[width=\linewidth]{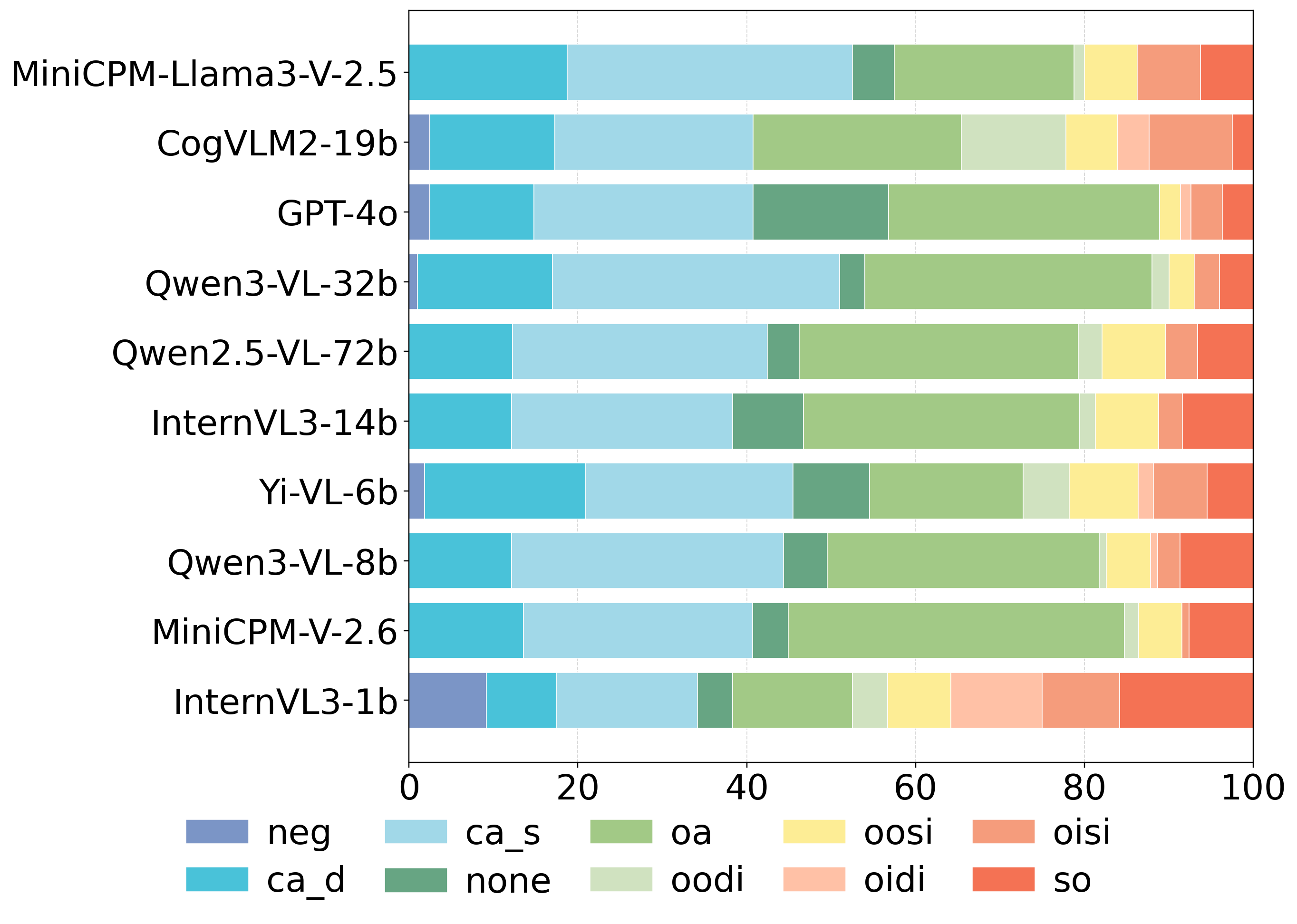}
        \footnotesize (b) Activity Description (abbreviated legend)
    \end{minipage}\hfill
    \begin{minipage}[!h]{0.5\textwidth}
        \centering
        \includegraphics[width=\linewidth]{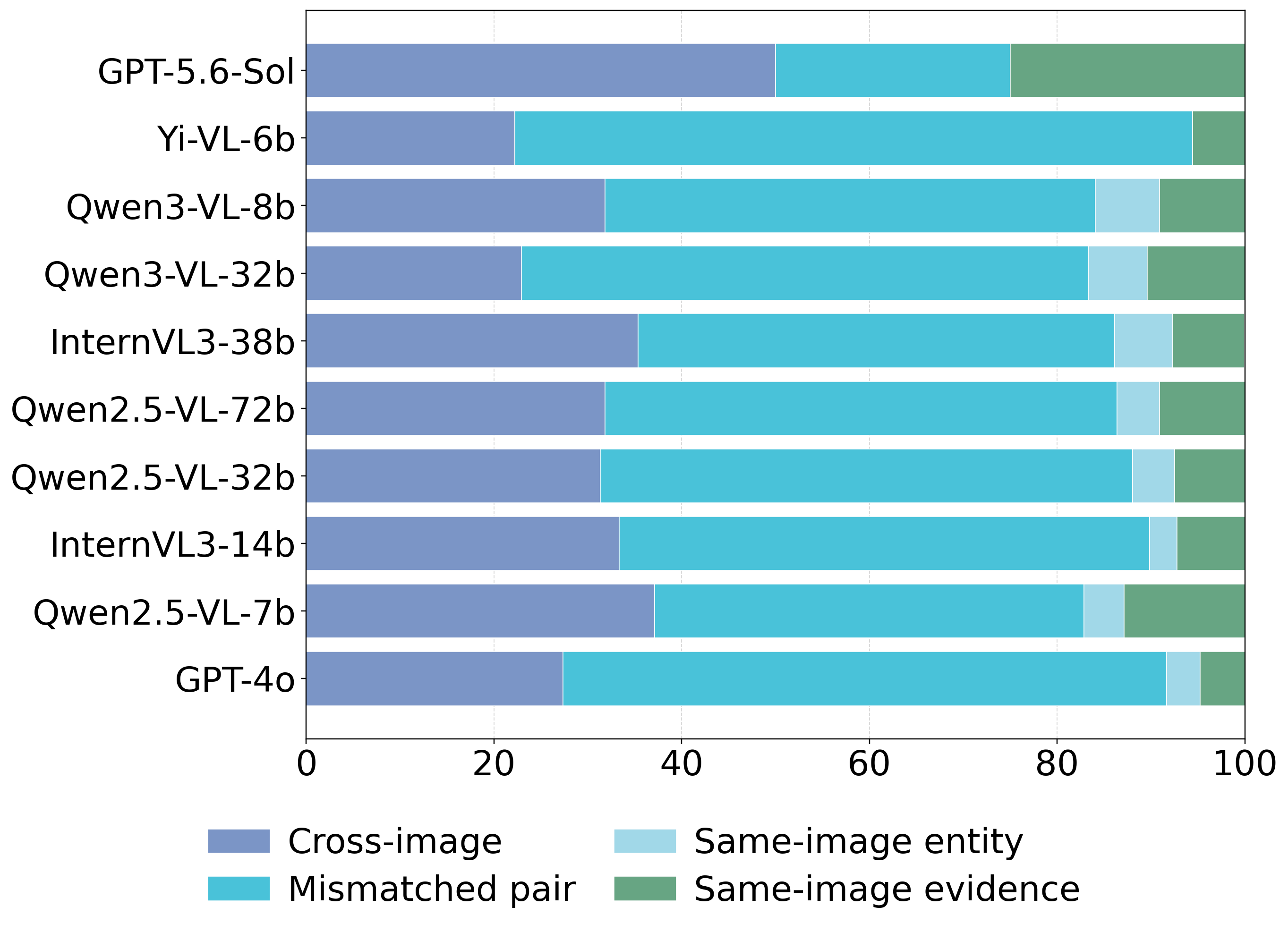}
        \footnotesize (c) Remote Interaction
    \end{minipage}
    \caption{Distributions of incorrect option or evidence-selection types across models for three tasks.}
    \label{fig:cross-task-errors}
\end{figure}
Figure~\ref{fig:external-correlations} shows that \textit{performance on existing benchmarks transfers unevenly to artistic educational imagery}. MMBench~\cite{liu2024mmbench} and AI2D~\cite{kembhavi2016ai2d} correlate strongly with several MUSE tasks, indicating partial overlap in perceptual and semantic capabilities, whereas BLINK~\cite{fu2024blink} exhibits inconsistent correlations across tasks. Activity Description and Relative Position assess capabilities underrepresented in existing benchmarks. Notably, AICA-Bench Emotion Reasoning~\cite{she2026aicabench} aligns moderately with MUSE's affective tasks, suggesting that emotion reasoning on conventional visual content only partially transfers to stylized expressions and implicit narratives in artworks. Overall, existing benchmarks explain only part of the model variation on MUSE, supporting its \textit{complementary coverage of artistic, affective, compositional, and cultural understanding}.

Figure~\ref{fig:cross-task-errors} reveals a common grounding failure across the three tasks. In Activity Localization, models usually select semantically relevant people or activities rather than random regions, but fail to identify the complete target extent. Activity Description (abbreviated legend labels are detailed in $\S$ \textsc{Question Generation}) errors similarly favor co-occurring or concatenated activities, indicating weak separation of the queried event from nearby visual semantics. In Remote Interaction, mismatched region-text pairs dominate, showing that models often accept plausible relations without verifying whether entities, regions, and evidence are jointly aligned. Overall, current VLMs capture \textit{coarse semantic relevance} but struggle with precise region-activity binding and image-specific relational grounding.
\subsection{Affective Computing Analysis}
\begin{figure}[t]
    \centering
    \includegraphics[width=0.8\linewidth]{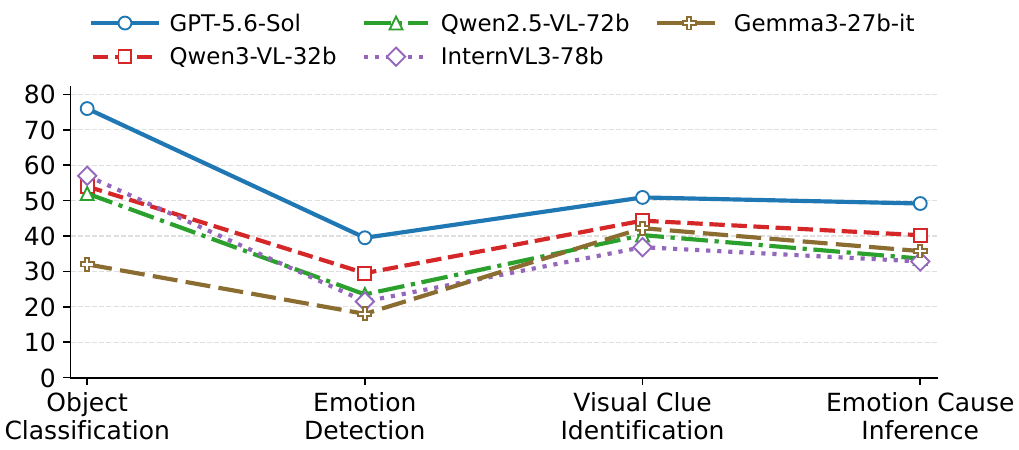}
    \caption{Five top large VLMs on affective computing.}
    \label{fig:emotion}
\end{figure}

Figure~\ref{fig:emotion} shows task-dependent ranking shifts across affective tasks, revealing that affective understanding is not a unified capability. Performance in character recognition or emotion classification does not reliably transfer to visual-evidence grounding or emotion-cause inference. Emotion Detection is the clearest bottleneck, reflecting the difficulty of interpreting stylized facial, bodily, and contextual cues. Despite differing metrics, within-task rankings indicate that \textit{current VLMs lack integrated affective reasoning from recognition to evidence and causal explanation}.

\begin{figure}[!h]
    \centering
    \begin{minipage}[c]{0.49\linewidth}
        \centering
        \includegraphics[width=0.95\linewidth]{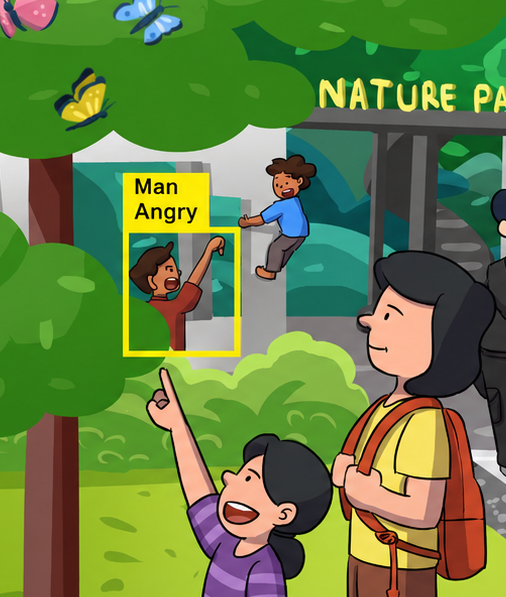}
        \footnotesize (a) Localization and emotion
    \end{minipage}\hfill
    \begin{minipage}[c]{0.49\linewidth}
        \centering
        \includegraphics[width=0.95\linewidth]{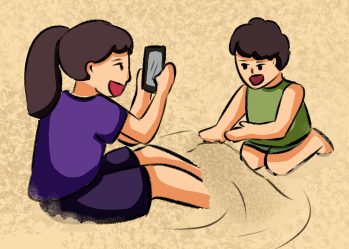}
        \footnotesize (b) Relative Position
    \end{minipage}
    \caption{Failure cases in Affective Computing and Relative Position.}
    \label{fig:failure-cases}
\end{figure}
\subsection{Visual Grounding is the Prerequisite of Accurate Affective Interpretation}
Figure~\ref{fig:failure-cases}(a) reveals a cascading failure across target grounding, affect recognition, and causal explanation. GPT-5.6-Sol correctly identifies the target man and attends to relevant cues, but misreads his stylized expression as \textit{Surprise}, indicating an affect-interpretation error rather than a grounding failure. Other models often shift attention to a salient child and predict \textit{Joy}, then justify the prediction using butterflies, birds, or nearby interactions. This suggests that \textit{errors in coordinate grounding and depth assignment leads models to construct a coherent explanation for the wrong character}. More broadly, flattened perspective and ambiguous occlusion in artistic images make affective reasoning depend on jointly resolving target identity, spatial structure, body posture, interactions, and scene context.

\subsection{Viewpoint-Aware Spatial Reasoning is a Persistent Bottleneck}
Figure~\ref{fig:failure-cases}(b) exposes a strong forced-relation bias in spatial reasoning. Although the ground truth specifies no definite lateral or vertical relation, 90.0\% and 73.3\% of models, respectively, predict one; depth reasoning is also unreliable, with only 43.3\% correctly identifying the girl as in front of the boy. No model resolves all three dimensions correctly. Current VLMs therefore oscillate between two failure modes: asserting \textit{definite relations under ambiguous evidence} or predicting \textit{None} across all dimensions and \textit{missing valid depth cues}. This reveals \textit{weak viewpoint-aware spatial reasoning} and \textit{poor calibration of spatial uncertainty}.

\section{Conclusion}
We introduced MUSE, a benchmark for evaluating multimodal understanding of artistic imagery in image-based language learning and educational interaction. Its construction framework decouples reusable visual-semantic annotations from task-specific question generation, enabling 12 tasks across five capability dimensions with control over question format and difficulty. Evaluation of 30 open-source and proprietary VLMs reveals task-dependent performance: models are reliable at scene and activity recognition but remain limited in visual grounding, affective interpretation, and compositional reasoning. Correlation analyses show that MUSE measures related yet non-redundant capabilities and complements general-purpose and emotion-reasoning benchmarks. Our error analyses identify recurring failures in precise region-activity binding, entity-evidence alignment, target grounding, and calibration under ambiguous spatial relations; these errors can propagate into coherent explanations for incorrectly grounded characters. Results indicate that scaling or stronger coarse recognition alone is insufficient. Reliable educational VLMs require region-aware grounding, integrated reasoning from perception to evidence and causes, and viewpoint-aware modeling of spatial uncertainty. MUSE provides a foundation for measuring progress toward these capabilities on artistic and culturally situated imagery.

\bibliographystyle{unsrtnat}
\bibliography{references}  






\appendix
\clearpage
\input{appendix}

\end{document}

%% file: appendix.tex
\section{Additional Analyses}

\subsection{Human-Model Comparison}

\begin{figure}[!ht]
    \centering
    \includegraphics[width=0.99\linewidth]{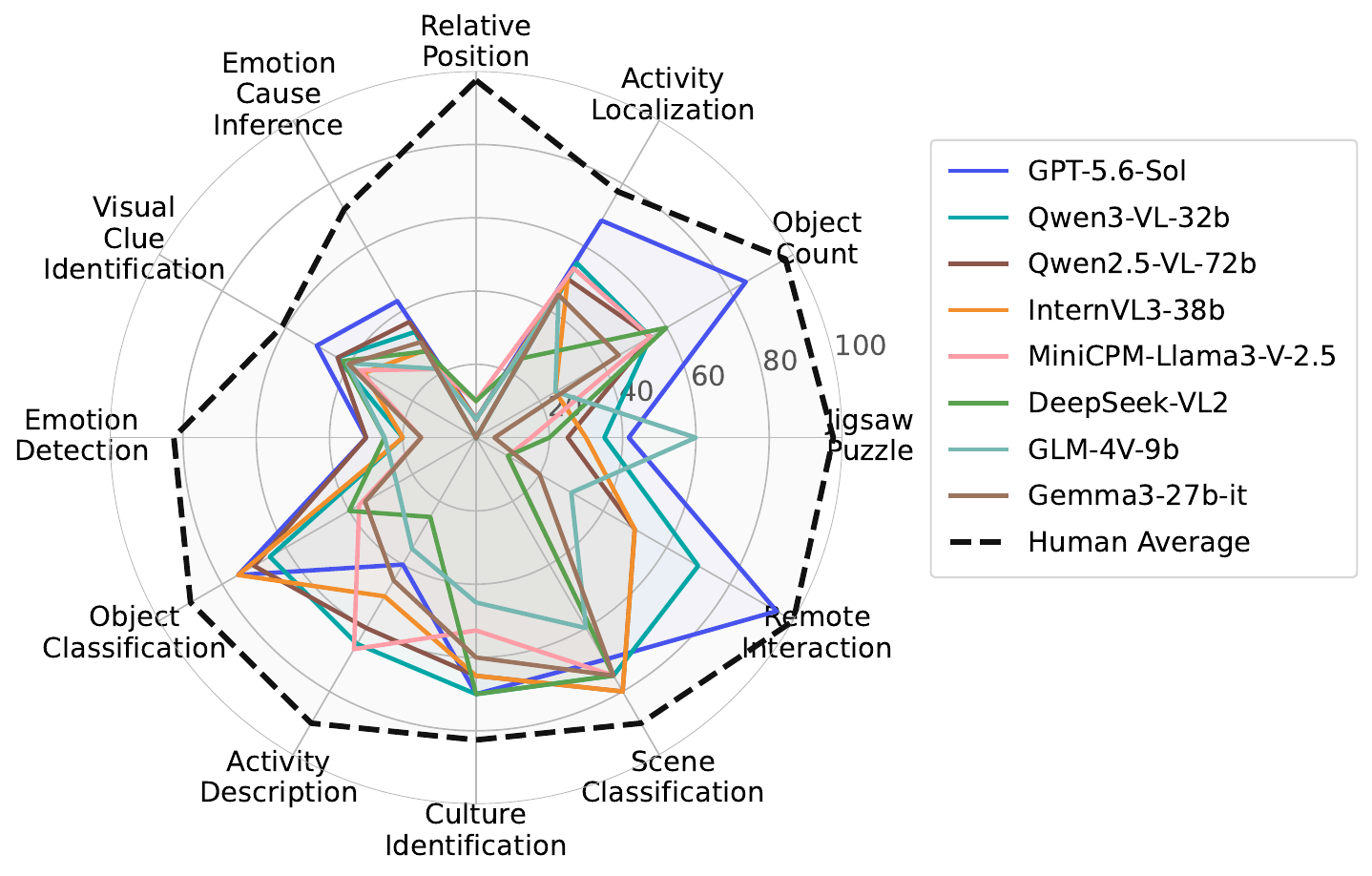}
    \caption{Comparison of human performance with top representatives from eight model families across the 12 MUSE tasks.}
    \label{fig:human result}
    
\end{figure}

To analyze the performance gap between humans and large VLMs across the 12 MUSE tasks, we randomly sampled 20 questions from each task and asked two annotators to answer them. We also collected the corresponding responses generated by different models for the same set of sampled questions.
Figure~\ref{fig:human result} shows that the human average forms the outer performance envelope on nearly all tasks, demonstrating a substantial gap between current VLMs and human multimodal understanding. The largest deficits occur in Relative Position, Emotion Detection, and Jigsaw Puzzle, where even the strongest models remain far below human performance. The gap is narrower for Activity Localization, Object Count, Remote Interaction, and Scene Classification, indicating stronger progress in visible-content recognition and selected relational tasks. Model profiles nevertheless vary considerably: GPT-5.6-Sol is strongest on Object Count and Remote Interaction, while GLM-4V-9b performs particularly well on Jigsaw Puzzle. These differences reinforce that no model family consistently approaches human performance across all capabilities.


\subsection{Performance across Taxonomy Dimensions}

\begin{figure}[!htbp]
    \centering
    \begin{minipage}[t]{0.75\linewidth}
        \centering
        \includegraphics[width=\linewidth]{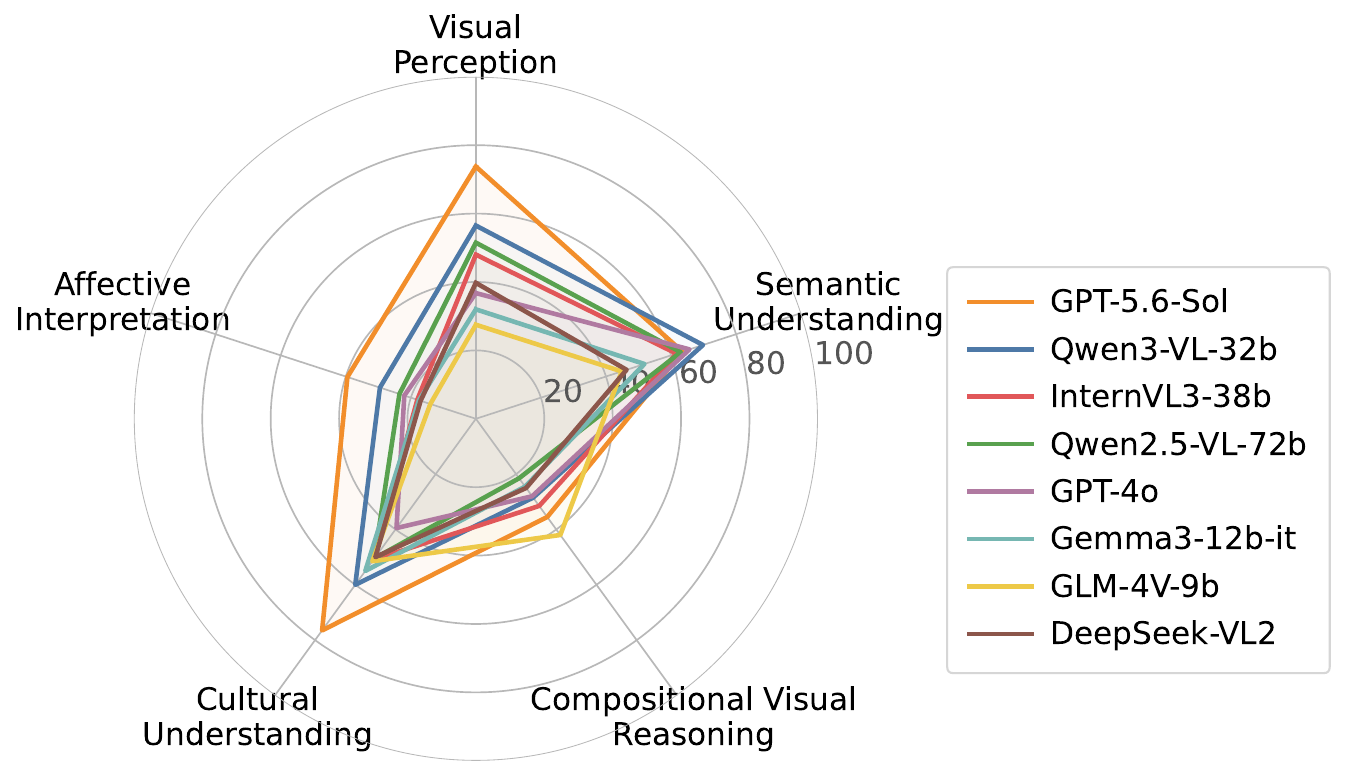}
        \small (a) Capability
    \end{minipage}\hfill
    \begin{minipage}[t]{0.75\linewidth}
        \centering
        \includegraphics[width=\linewidth]{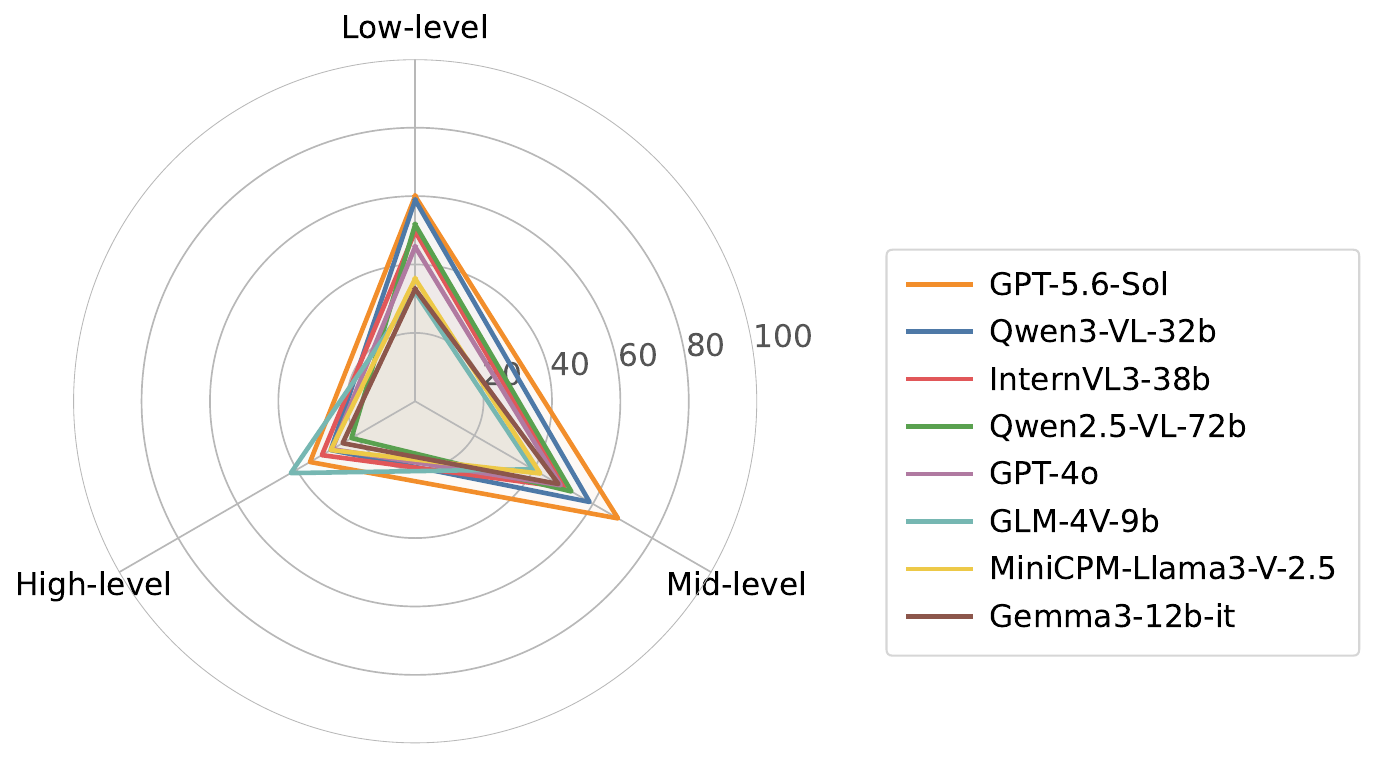}
        \small (b) Difficulty
    \end{minipage}\hfill
    \begin{minipage}[t]{0.75\linewidth}
        \centering
        \includegraphics[width=\linewidth]{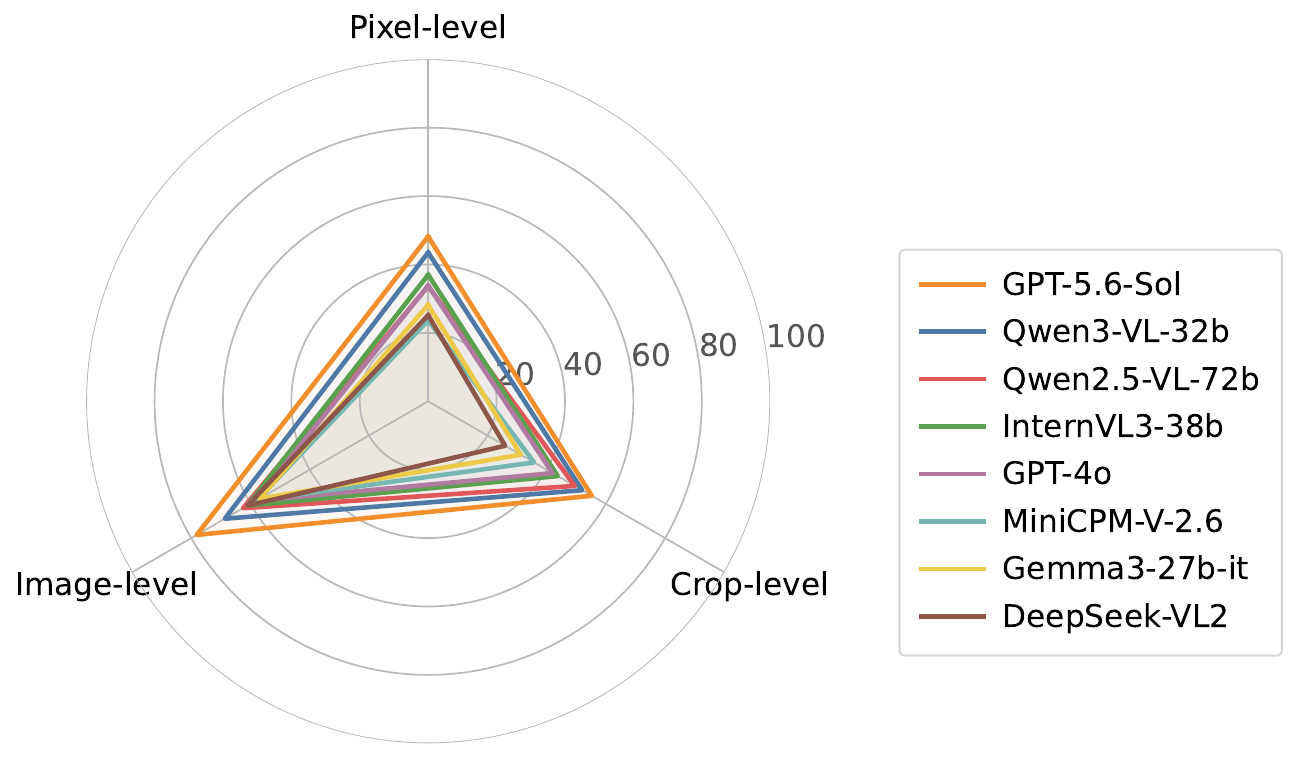}
        \small (c) Granularity
    \end{minipage}
    \caption{Top representatives from eight model families across capability, difficulty, and granularity dimensions.}
    \label{fig:taxonomy radars}
\end{figure}

Figure~\ref{fig:taxonomy radars} reveals consistent performance imbalances across the MUSE taxonomy. GPT-5.6-Sol has the strongest and most balanced overall profile, although other models retain dimension-specific advantages. Across capabilities, semantic and cultural understanding are generally stronger than affective interpretation and compositional visual reasoning. Performance also tends to decrease from low- and mid-level tasks to high-level reasoning, showing that success on recognition and semantic perception does not reliably extend to more complex inference. Across spatial granularities, image-level understanding is consistently strongest, whereas pixel-level understanding is weakest and crop-level performance remains intermediate. This pattern indicates that global scene interpretation is more mature than precise local grounding.


\subsection{Invalid Response Analysis}

\begin{figure}[!ht]
    \centering
    \includegraphics[width=0.99\textwidth]{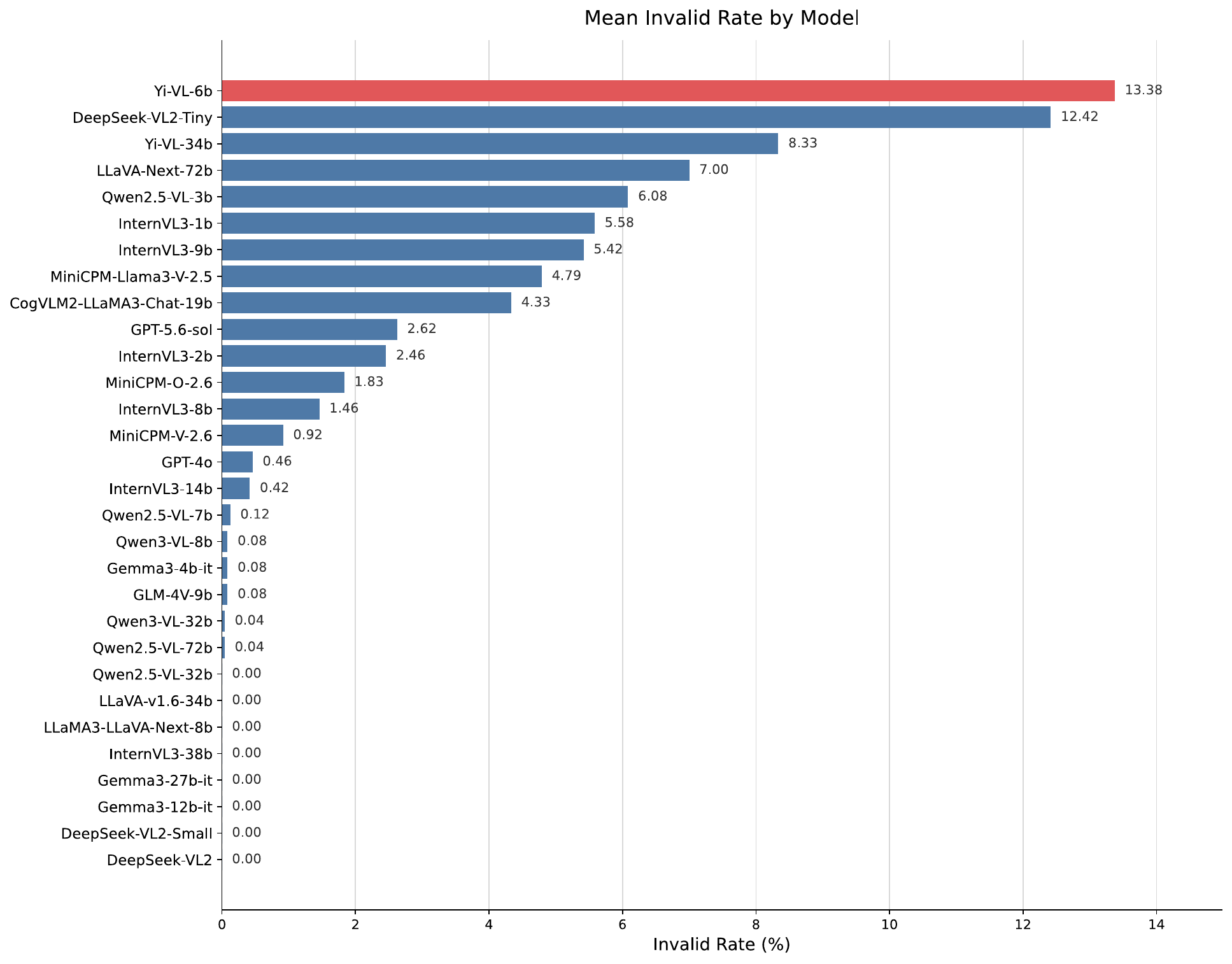}
    \caption{Mean invalid response rate for each VLM.}
    \label{fig:invalid rate}
    
\end{figure}

Figure~\ref{fig:invalid rate} shows a highly skewed distribution of invalid responses. Most models have invalid rates below 1\%, and several produce no invalid responses. In contrast, Yi-VL-6b and DeepSeek-VL2-Tiny exceed 12\%, while Yi-VL-34b, LLaVA-NeXT-72b, Qwen2.5-VL-3b, and smaller InternVL3 variants also exhibit elevated rates. Invalid responses are not determined solely by model scale: models within the same family vary substantially, and GPT-5.6-Sol retains a 2.62\% invalid rate despite its strong task performance. Thus, response-format reliability constitutes a distinct evaluation concern alongside answer correctness.


\section{Examples for Selected Tasks}
\subsection{Cultural Identity} 
\begin{figure}[!ht]
    \centering
    \includegraphics[width=\linewidth]{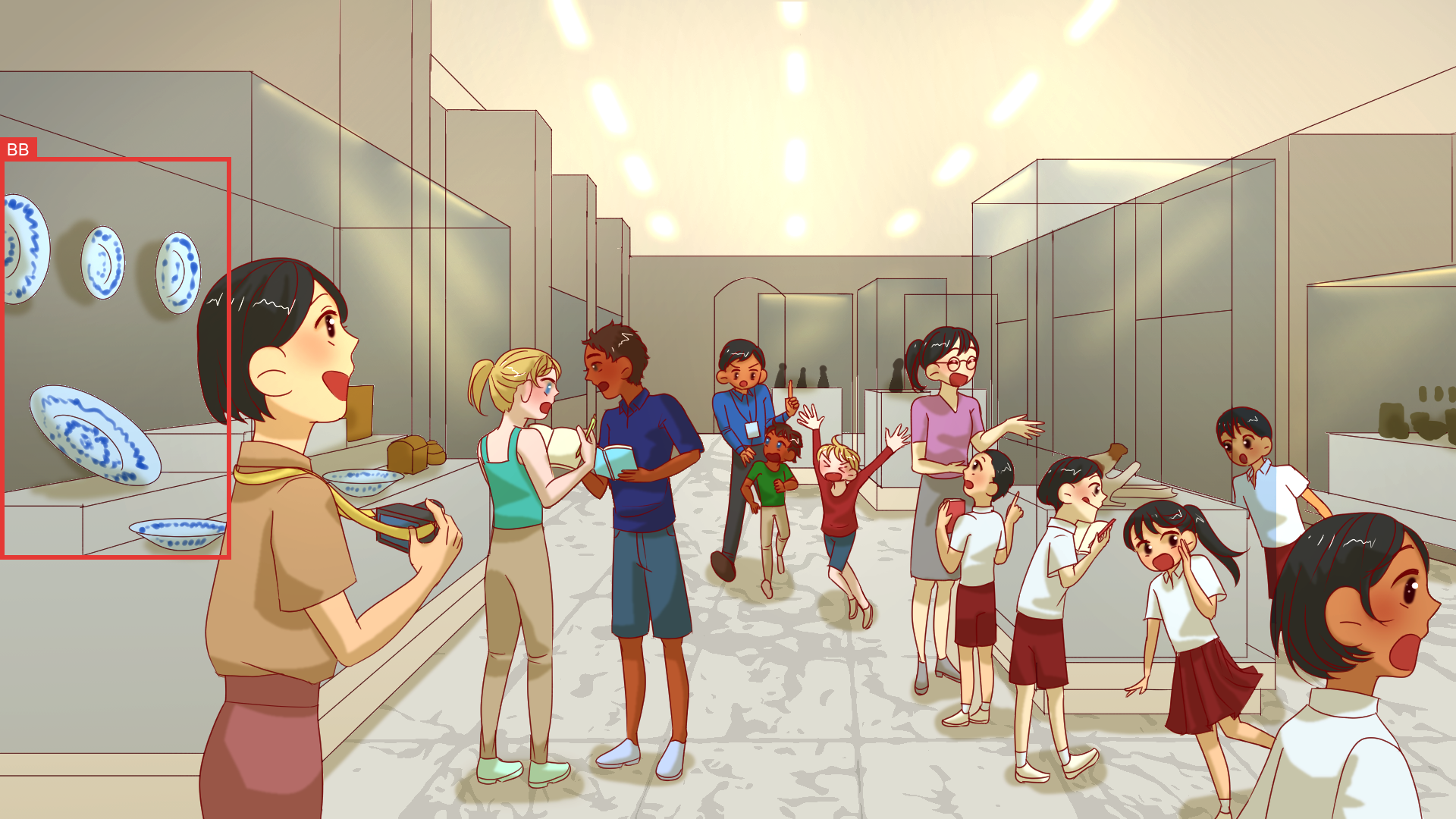}
    \caption{Example image for Cultural Identitfication.}
    \label{fig:example cultural}
    
\end{figure}

The following is an example Cultural Identification question and its corresponding image Figure~\ref{fig:example cultural}. The red bounding box is included solely to facilitate interpretation and is not shown to the VLMs during inference.


\noindent\textbf{Question}

Given an image, identify the culture that is most relevant to the content within the bounding box [1.0945860806163514e-17, 0.19206680584551108, 0.15845070422535204, 0.4906054279749479]. The bounding box coordinates are in COCO-format [xmin, ymin, width, height]. All the coordinates are in percentages between 0 to 1.
Please select the most appropriate culture option from the following options.

\noindent options:

A. Western
B. China
C. Europe
D. Muslim

\noindent The response should be in the following format.
The answer should be A / B / C / D only.

\noindent Constraints:

- Do not include any additional text or explanation.

\subsection{Jigsaw Puzzle}
\begin{figure}[!ht]
    \centering
    \includegraphics[width=\linewidth]{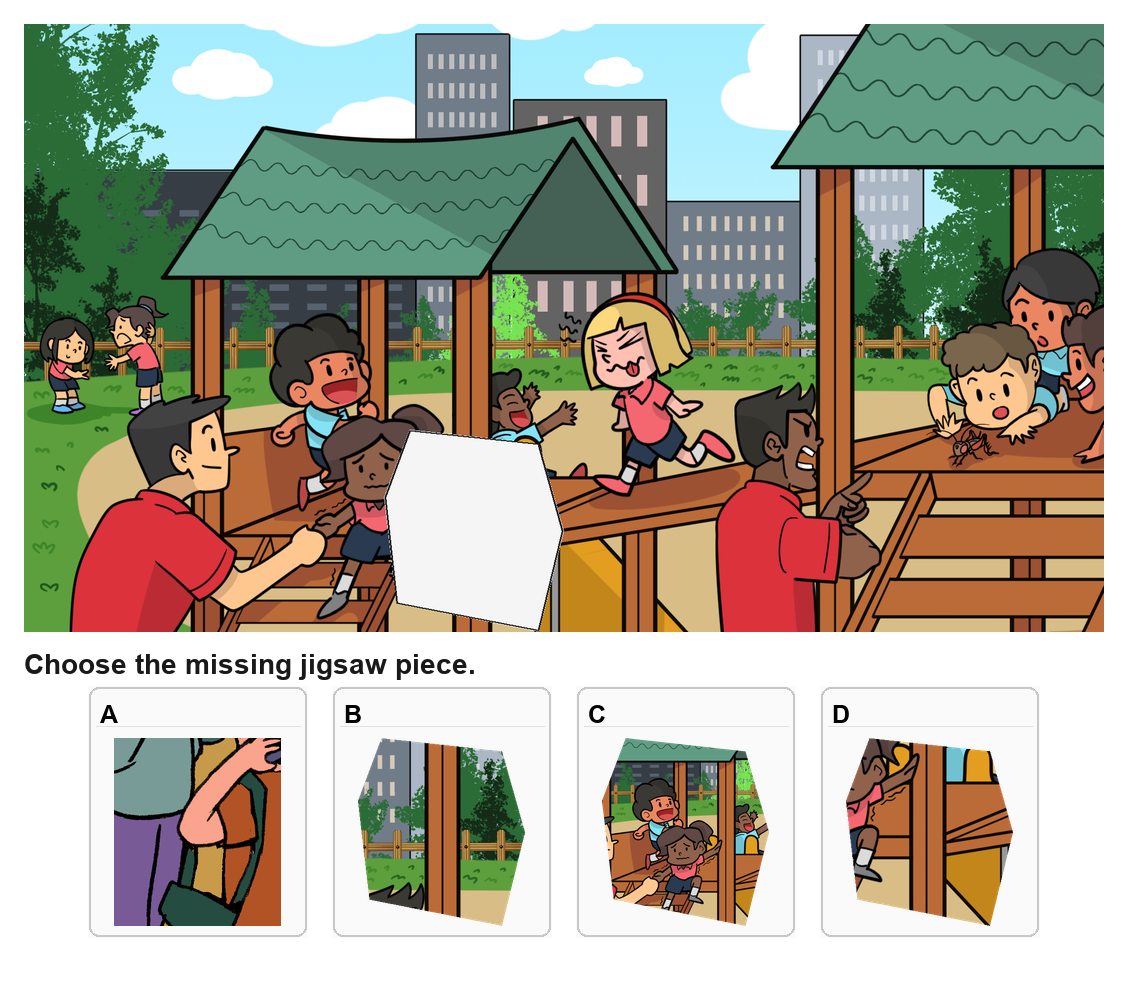}
    \caption{Example image for Jigsaw Puzzle.}
    \label{fig:example jigsaw}
    
\end{figure}

The following is an example Jigsaw Puzzle question and its corresponding image Figure~\ref{fig:example jigsaw}.

\FloatBarrier

\noindent\textbf{Question}

Given an image with a missing region, select the one candidate image piece that best completes the image.

\noindent Options:
A
B
C
D

\noindent Instructions:

1. Exactly one option is correct.

2. Answer using only a single uppercase letter: A, B, C, or D.

3. Do not output any explanation, reasoning, punctuation, or additional text.

\subsection{Affective Computing}
The following shows a four-turn-sequence questions for Object Classificaiton, Emotion Detection, Visual Cause Indentification, and Emotion Cause Inference.

\begin{figure}[h]
    \centering
    \includegraphics[width=\linewidth]{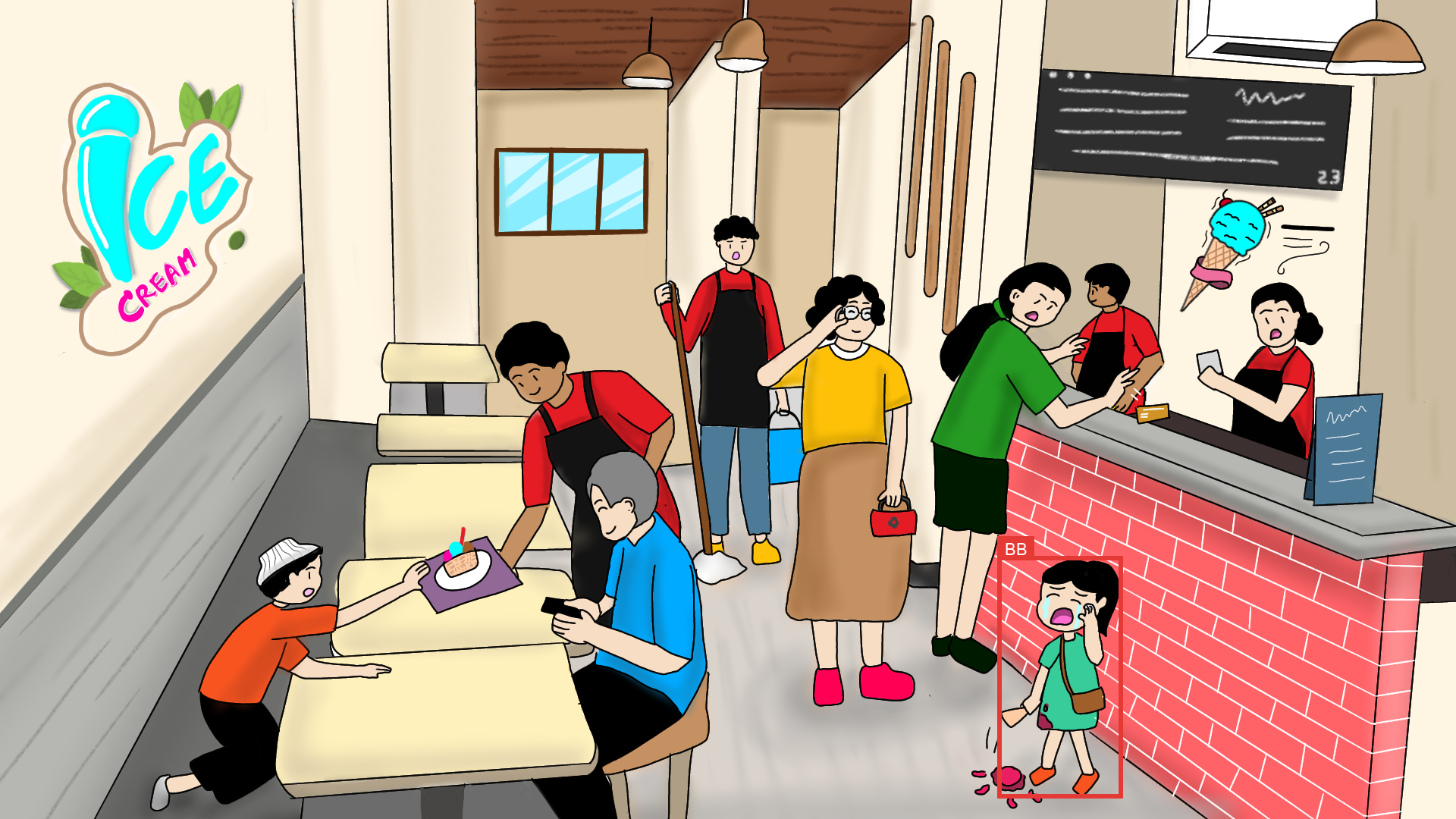}
    \caption{Example image for affective computing.}
    \label{fig:example emotion}
    
\end{figure}

\textbf{Question - Object Classification}

Given an image and a bounding box, identify the object category corresponding to the bounding box. The bounding box coordinates are in COCO-format [xmin, ymin, width, height], with all values between 0 and 1.

\noindent[bounding box] [0.685, 0.679, 0.086, 0.295]

\noindent[object options] 
A. Boy
B. Woman
C. Baby
D. Girl
E. Man

\noindent Return exactly one line in this format:

\noindent [option] <selected object option letter>

\noindent Constraints:

- Output must start with [option]

- Followed by a space and a single uppercase letter (A–Z)

- Do not include any additional text or explanation.

\noindent\textbf{Question - Emotion Detection}

Given the same image and bounding box, identify the emotion of the person inside the bounding box.

\noindent[bounding box] [0.685, 0.679, 0.086, 0.295]

\noindent[emotion options]

A. Guilt
B. Confusion
C. Sadness
D. Neutral
E. Boredom
F. Disgust
G. Surprise
H. Joy
I. Anger
J. Anxiety
K. Fear

Return exactly one line in this format:

\noindent[emotion] <selected emotion option letter>

\noindent Constraints:

- Output must start with [emotion]

- Followed by a space and a single uppercase letter (A–Z)

- Do not include any additional text or explanation.

\noindent\textbf{Question - Visual Clue Identification}

Based on the image and bounding box below, describe the observable visual clues that support the previously identified emotion.

\noindent[bounding box] [0.685, 0.679, 0.086, 0.295]

\noindent[emotion] \{identified$\_$emotion\}

Return the result in the following format.

\noindent [visual clues] <identified visual clues>

\noindent\textbf{Question - Emotion Cause}

Based on the image, the bounding box, and the visual clues above, infer the most likely cause of the identified emotion.

Return the result in the following format.

\noindent[emotion cause] <inferred emotion cause>

\section{Model Hyperparameters}

Table~\ref{tab:model_precision} summarizes the computation dtypes used during inference. Most evaluated model families use BF16, while the LLaVA-NeXT models use FP16. CogVLM2 uses BF16 when supported by the hardware and otherwise falls back to FP16. We retain the default dtypes specified by the corresponding inference scripts to reflect standard deployment settings and apply the same numerical configuration across all MUSE tasks for each model.

\begin{table}[t]
\centering
\setlength{\tabcolsep}{30pt}
\begin{tabular}{llc}
\toprule
\textbf{Family} & \textbf{Model} & \textbf{Dtype} \\
\midrule
CogVLM2
& \texttt{cogvlm2-llama3-chat-19b}
& BF16$^{\dagger}$ \\

\midrule
\multirow{3}{*}{DeepSeek-VL2}
& \texttt{deepseek-vl2-tiny}  & BF16 \\
& \texttt{deepseek-vl2-small} & BF16 \\
& \texttt{deepseek-vl2}       & BF16 \\

\midrule
\multirow{3}{*}{Gemma-3}
& \texttt{gemma-3-4b-it}  & BF16 \\
& \texttt{gemma-3-12b-it} & BF16 \\
& \texttt{gemma-3-27b-it} & BF16 \\

\midrule
GLM-4V
& \texttt{glm-4v-9b} & BF16 \\

\midrule
\multirow{6}{*}{InternVL3}
& \texttt{internvl3-1b}  & BF16 \\
& \texttt{internvl3-2b}  & BF16 \\
& \texttt{internvl3-8b}  & BF16 \\
& \texttt{internvl3-9b}  & BF16 \\
& \texttt{internvl3-14b} & BF16 \\
& \texttt{internvl3-38b} & BF16 \\

\midrule
\multirow{6}{*}{LLaVA-NeXT}
& \texttt{llama3-llava-next-8b}       & FP16 \\
& \texttt{llava-next-72b-hf}          & FP16 \\
& \texttt{llava-v1.6-34b-hf}          & FP16 \\
& \texttt{llava-v1.6-mistral-7b-hf}   & FP16 \\
& \texttt{llava-v1.6-vicuna-7b-hf}    & FP16 \\
& \texttt{llava-v1.6-vicuna-13b-hf}   & FP16 \\

\midrule
\multirow{3}{*}{MiniCPM}
& \texttt{minicpm-llama3-v-2\_5} & BF16 \\
& \texttt{minicpm-o-2\_6}        & BF16 \\
& \texttt{minicpm-v-2\_6}        & BF16 \\

\midrule
\multirow{4}{*}{Qwen2.5-VL}
& \texttt{qwen2\_5\_vl\_3b}  & BF16 \\
& \texttt{qwen2\_5\_vl\_7b}  & BF16 \\
& \texttt{qwen2\_5\_vl\_32b} & BF16 \\
& \texttt{qwen2\_5\_vl\_72b} & BF16 \\

\midrule
\multirow{2}{*}{Qwen3-VL}
& \texttt{qwen3\_vl\_8b-instruct}  & BF16 \\
& \texttt{qwen3\_vl\_32b-instruct} & BF16 \\

\midrule
\multirow{2}{*}{Yi-VL}
& \texttt{yi-vl-6b}  & BF16 \\
& \texttt{yi-vl-34b} & BF16 \\
\bottomrule
\end{tabular}

\begin{minipage}{\linewidth}
\footnotesize
$^{\dagger}$The CogVLM2 script uses BF16 when supported by the hardware and otherwise falls back to FP16.
\end{minipage}
\caption{Default computation dtypes used by the inference scripts.}
\label{tab:model_precision}
\end{table}

Table~\ref{tab:generation_settings} summarizes the default generation configuration used in our inference pipeline. We disable sampling to obtain deterministic outputs and set \texttt{max\_new\_tokens} to 1024 to accommodate both short structured answers and open-ended responses. Consequently, temperature, top-$k$, and top-$p$ do not affect decoding. All other unspecified parameters inherit the corresponding model or library defaults, preserving each model's native beam-search, repetition-control, and caching behavior.

\begin{table}[!htbp]
\centering

\setlength{\tabcolsep}{30pt}
\begin{tabular}{lll}
\toprule
\textbf{Parameter} & \textbf{Default} & \textbf{Effect under Default Setting} \\
\midrule
\texttt{max\_new\_tokens}       & 1024          & Maximum generated length \\
\texttt{do\_sample}             & \texttt{False} & Deterministic decoding \\
\texttt{temperature}            & \texttt{None}  & Inactive without sampling \\
\texttt{top\_k}                 & \texttt{None}  & Inactive without sampling \\
\texttt{top\_p}                 & \texttt{None}  & Inactive without sampling \\
\texttt{num\_beams}             & \texttt{None}  & Uses the library default \\
\texttt{repetition\_penalty}    & \texttt{None}  & Uses the library default \\
\texttt{num\_return\_sequences} & \texttt{None}  & Uses the library default \\
\texttt{use\_cache}             & \texttt{None}  & Uses the model default \\
\texttt{cache\_implementation}  & \texttt{None}  & Uses the model default \\
\bottomrule
\end{tabular}
\caption{Default generation settings used in our inference pipeline. Parameters set to \texttt{None} use the underlying model or library defaults. Since sampling is disabled, sampling-specific parameters such as temperature, top-$p$, and top-$k$ are inactive under the default configuration.}
\label{tab:generation_settings}
\end{table}


\section{Annotation Process}

\paragraph{Annotator Recruitment and Preparation.}
We recruited 127 undergraduate and postgraduate student annotators. Before annotation, they completed a 0.5-hour training session based on written guidelines specifying annotation categories, bounding-box conventions, and procedures for resolving ambiguous artistic content. Feedback from a pilot annotation stage was incorporated to further clarify the guidelines.


\paragraph{Time, Compensation, and Cost.}
The average cost of commissioning each image from freelance artists was approximately USD~36. Annotators spent approximately 3-4 minutes per image, which varies based on task categories, corresponding to 4 hours of annotation. They were compensated at USD~16 per hour. Additional costs included platform fees . Compensation was set with reference to local institutional policy.

\paragraph{Ethical and Data-Handling Considerations.}
Annotators were informed about the purpose and intended use of the dataset. We collected no personal information beyond what was necessary for compensation and quality control. Potentially sensitive cultural or affective annotations were reviewed carefully.